\documentclass{article}

\usepackage{PRIMEarxiv}

\usepackage[utf8]{inputenc} 
\usepackage[T1]{fontenc}    
\usepackage{hyperref}       
\usepackage{url}            
\usepackage{booktabs}       
\usepackage{nicefrac}       
\usepackage{microtype}      
\usepackage{lipsum}
\usepackage{fancyhdr}       
\usepackage{graphicx}       
\graphicspath{{media/}}     

\usepackage{xcolor}         

\usepackage{subcaption}
\usepackage{graphicx}
\usepackage{wrapfig}
\usepackage{float}
\usepackage{svg}
\usepackage{algorithm}
\usepackage{multicol}
\usepackage{algpseudocode}
\usepackage{ulem}    
\usepackage{paralist, tabularx}
\usepackage{flafter}
\usepackage{icomma}
\usepackage{amsmath}
\usepackage{amsfonts}
\usepackage{bm}

\def\vx{{\bm{x}}}

\DeclareMathOperator*{\argmax}{arg\,max}

\title{A Principled Evaluation Protocol for Comparative Investigation of the Effectiveness of DNN Classification Models on Similar-but-non-identical Datasets
}

\author{
  Esla Timothy Anzaku\,\thanks{Corresponding author: \texttt{eslatimothy.anzaku@ugent.be}} \,
  \thanks{Department of Electronics and Information Systems, Ghent University, Belgium} \,
  \thanks{Center for Biosystems and Biotech Data Science, Ghent University Global Campus, Republic of Korea}
   \And Haohan Wang \thanks{School of Information Sciences, University of Illinois Urbana-Champaign, USA}
   \And Arnout Van Messem \thanks{Department of Mathematics, University of Li\`ege, Belgium}
   \And Wesley De Neve \footnotemark[2] \,  \footnotemark[3]
   \newline
}

\begin{document}
\maketitle

\begin{abstract}
  Deep Neural Network (DNN) models are increasingly evaluated using new replication test datasets, which have been carefully created to be similar to older and popular benchmark datasets. However, running counter to expectations, DNN classification models show significant, consistent, and largely unexplained degradation in accuracy on these replication test datasets. While the popular evaluation approach is to assess the accuracy of a model by making use of all the datapoints available in the respective test datasets, we argue that doing so hinders us from adequately capturing the behavior of DNN models and from having realistic expectations about their accuracy. Therefore, we propose a principled evaluation protocol that is suitable for performing comparative investigations of the accuracy of a DNN model on multiple test datasets, leveraging subsets of datapoints that can be selected using different criteria, including uncertainty-related information. By making use of this new evaluation protocol, we determined the accuracy of $564$ DNN models on both (1) the CIFAR-10 and ImageNet datasets and (2) their replication datasets. Our experimental results indicate that the observed accuracy degradation between established benchmark datasets and their replications is consistently lower (that is, models do perform better on the replication test datasets) than the accuracy degradation reported in published works, with these published works relying on conventional evaluation approaches that do not utilize uncertainty-related information.
\end{abstract}

\section{Introduction}
\label{section:introduction}

The purpose of evaluating trained machine learning models on held-out test datasets is to obtain unbiased estimates of their performance on datapoints that were not used for training purposes. A newer assessment approach in computer vision is to evaluate models on independent test datasets~\cite{Recht_Do_ImageNet, Lu_Harder_or} that have been created by carefully replicating the way older and highly popular datasets were constructed. We expect the accuracy on the replicated datasets to be similar to the accuracy on the original datasets. However, in reality, models suffer from a significant and consistent degradation in accuracy when evaluation is done using replicated datasets. For example,~\cite{Recht_Do_ImageNet} created two test datasets, namely CIFAR-10.1 and ImageNetV2, by carefully following the way the CIFAR-10~\cite{Cifar10} and ImageNet~\cite{Deng_ImageNet} datasets were constructed, respectively. They subsequently evaluated a large number of models on these test datasets and reported consistent and significant accuracy drops: $3\% - 15\%$ on CIFAR-10.1 and $11\% - 14\%$ on ImageNetV2. For convenience, in the remainder of this paper, we will refer to the original validation dataset partition of the ImageNet dataset as ImageNetV1.

Even though care is taken to ensure that the replicated datasets closely match their original datasets by design, it is reasonable to expect some characteristics of these datasets to still differ. Examples of factors that could account for such characteristic differences include: the under-representation of some sub-populations, unintended differences in sampling strategies~\cite{Antonio_Unbiased_Look_At}, variation in the ratio of easy datapoints to difficult datapoints~\cite{ Recht_Do_ImageNet, Lu_Harder_or}, different degrees of label errors~\cite{Northcutt_Pervasive_Label}, and statistical bias during the replication process~\cite{Engstrom_Identifying_Statistical_2020}. We conjecture that the degradation in the accuracy of a model on similar-but-non-identical test datasets does not necessarily indicate that the model fails to generalize well outside the training environment. Instead, there is a possibility that the adopted evaluation approaches do not adequately capture the rich information that DNN classification models generate.

DNN classification models generally produce a softmax vector. From this vector, we can extract two outputs for any given input datapoint -- the predicted label and the corresponding predicted probability\footnote{In this paper, we refer to the maximum value of the softmax vector as the predicted probability or confidence in an interchangeable manner.}. The conventional approach for evaluating the accuracy of a model does not take into account the probabilities generated by the model, thereby treating every prediction equally. In this paper (see Section~\ref{section:proposed_evaluation_protocol}), we propose a principled evaluation protocol to assess the accuracy of a model on multiple similar-but-non-identical test datasets.

Consider a scenario where we seek to compare the accuracy of a model on two similar-but-non-identical test datasets. Instead of determining the accuracy using all the datapoints available in each dataset, we first create two subsets (one for each dataset), such that these two subsets match according to a particular matching criterion. We then evaluate the accuracy of the model on these two matched subsets. In this paper, the different matching criteria we study are ($i$) matching by both the predicted labels and their corresponding predicted probabilities and ($ii$) matching by the generated model probabilities.

\begin{wrapfigure}{r}{0.65\textwidth}
\centering
    \includegraphics[width=0.65\textwidth]{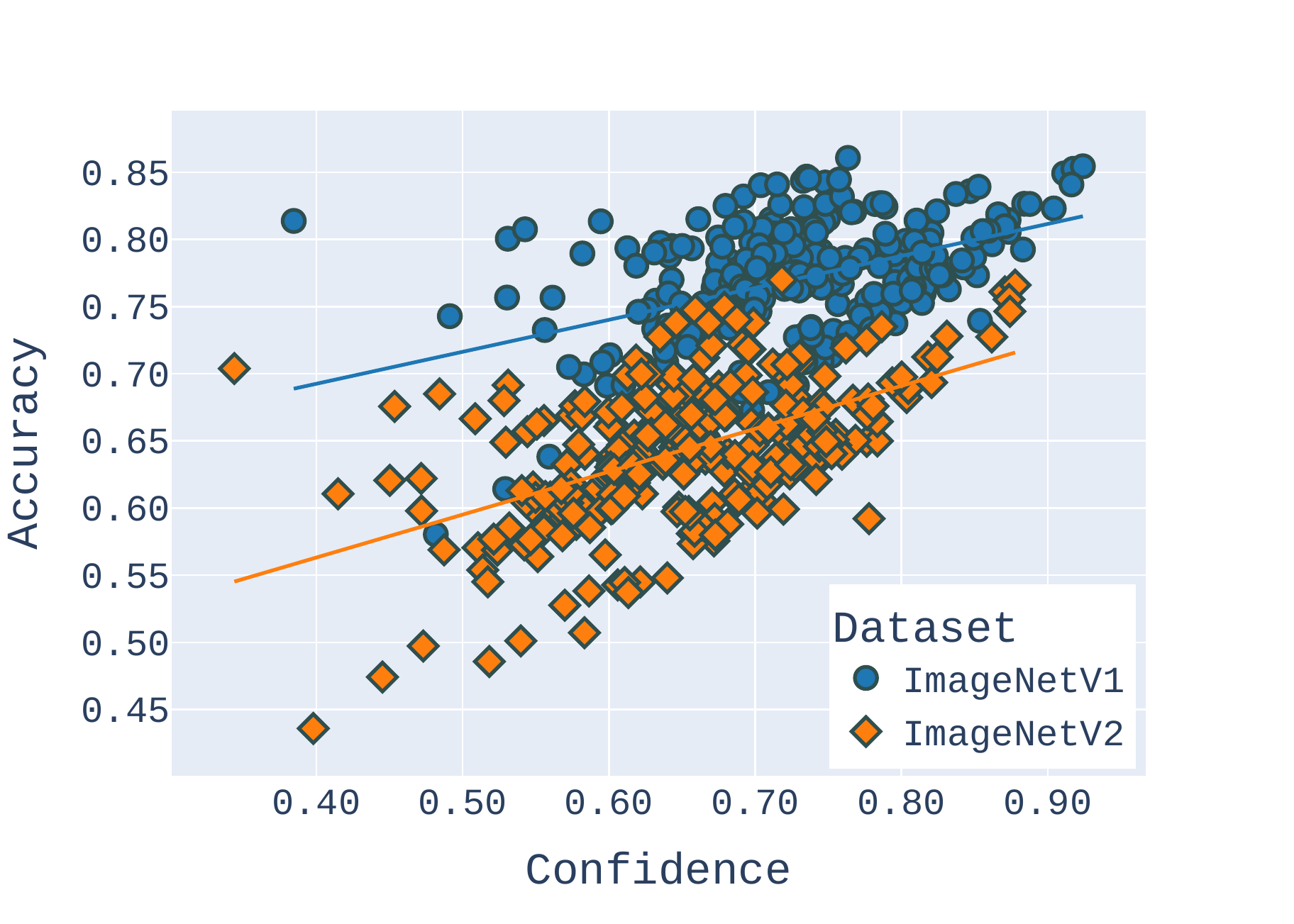}
    \caption{Scatter plot showing accuracy versus confidence for $286$ pre-trained ImageNet models. Each point corresponds to the accuracy and the confidence of a model on either the ImageNetV1 or ImageNetV2 dataset, with confidence denoting the average of the predicted probabilities for all datapoints.}
    \label{fig:scatter_plots_acc_vs_conf_imagenet}
\end{wrapfigure}

Although it has been reported that the softmax probabilities generated by DNN models are usually over-confident~\cite{Guo_on_calibration_of_2017, hein_why_relu_networks_2019}, there are research efforts~\cite{hendrycks_a_baseline_for_2019, mukhoti_calibrating_deep_2020, pearce_understanding_softmax_confidence_2021} that acknowledge that softmax probabilities can serve as a baseline for the uncertainty estimates that models make about their predictions. We argue that these softmax probabilities can provide additional insight into the behavior of models that the use of zero-one loss or accuracy alone cannot capture; one such aspect is the relationship between the accuracy and the uncertainty of a model. A property of DNN models that is of high interest to practitioners is for a DNN model to be more correct when more certain and to be less correct when less certain. To observe the relationship between accuracy and confidence, we evaluated the accuracy and confidence of $286$ pre-trained ImageNet models on the ImageNetV1 and ImagenetV2 datasets. Figure~\ref{fig:scatter_plots_acc_vs_conf_imagenet}
summarizes the obtained results by means of a scatter plot. Here, the confidence of a model is calculated as the average of the predicted probabilities associated with the predicted labels, for all the datapoints in the test dataset of interest. In Figure~\ref{fig:scatter_plots_acc_vs_conf_imagenet}, we can observe that models generally tend to be less confident and less accurate on the ImageNetV2 dataset, whereas models tend to be more confident and more accurate on the original ImageNet validation set (ImageNetV1).

The aforementioned observation brings up interesting questions that warrant further investigation. Are the models signalling that the characteristics of the two datasets differ in more ways than we can explicitly model? Are pre-trained models too sensitive to their training settings which require that the training and test environments are independently and identically distributed ($i.i.d.$)? Using evaluation protocols that adequately capture the behavior of DNN models is a step forward towards finding answers to this kind of questions. The evaluation protocol that we introduce in this paper implicitly assesses how consistent the accuracy metric is with respect to the uncertainties generated by DNN classification models. For example, if we sample subsets of predictions with similar predicted probabilities (matched by predicted probabilities) from two test datasets, how large will the accuracy gap between these subsets be? Intuitively, we expect such a gap to be narrow ($i$) if the probabilities carry important information about the uncertainty (or confidence) that the model has on its predictions and ($ii$) if the uncertainty information correlates well with the accuracy metric. Indeed, our extensive evaluations of $278$ CIFAR-10 models and $286$ ImageNet models, as presented in Section~\ref{section:experimental_setup_and_results}, provide compelling evidence for the utility of our evaluation approach. In particular, if we leverage the uncertainty-related information generated by models, we find that the degradation in accuracy across all evaluated models may not be as large as has been previously reported.

The contributions of this paper are as follows: 
\begin{compactenum}
    \item we draw attention to a weakness when comparing the accuracy of models on multiple similar-but-non-identical datasets;
    \item we propose a principled evaluation protocol for investigating the accuracy of models on multiple test datasets, with this evaluation protocol taking into account both the predicted labels and the predicted probabilities that are generated by models, thus allowing for more nuanced interpretation of model behavior; and
    \item by evaluating $564$ pre-trained DNN models, we provide empirical evidence that indicates that the deterioration in accuracy may not be as bad as has been reported before, if we leverage the uncertainty-related information generated by DNN models.
\end{compactenum}

In the remainder of this paper, we describe our principled evaluation protocol in detail in Section~\ref{section:proposed_evaluation_protocol}. Furthermore, we summarize related research efforts in Section~\ref{section:related_research}. Next, we discuss our experimental setup and our experimental results in Section~\ref{section:experimental_setup_and_results}. Finally, we present our conclusions and a number of directions for future research in Section~\ref{section:conclusion_and_future_work}. 

\section{Proposed Evaluation Protocol}
\label{section:proposed_evaluation_protocol}

\subsection{Background}

In this section, we describe our principled evaluation protocol, which enables us to gain more insight into the performance of models when evaluation is done using multiple datasets. Although the newly introduced protocol could be used to evaluate any supervised classification model that generates predicted labels and their corresponding probabilities, in this paper, we restrict our scope to supervised DNN classification models for computer vision. In machine learning, supervised classification involves observing several examples of a random vector $\bm{X}$ and an associated random variable $Y$, with the aim of learning to predict $Y$ from $\bm{X}$. This is usually done by estimating the conditional probability distribution $P(Y | \bm{X})$~\cite{Goodfellow_Deep_Learning_Book_2016}. The formulation of our evaluation protocol assumes that we are provided with:
\begin{compactitem}
    \item an $i.i.d.$ dataset $\mathcal{D} = \{ (\displaystyle \vx^{(i)}, y^{(i)})\}_{i=1}^{N} \subset \mathcal{X \times Y}$, which is further partitioned into disjoint train ($\mathcal{D_{\mathrm{train}}}$), validation ($\mathcal{D_{\mathrm{val}}}$), and test ($\mathcal{D_{\mathrm{test1}}}$)\footnote{The ImageNet dataset does not include the ground truth labels for the test partition. Therefore, the accuracy on the validation partition is usually reported in published articles.} datasets, where $\mathcal{X}$ and $\mathcal{Y} = \{1, 2, ..., K\}$ denote the input space and label space, respectively, and where $K$ represents the number of classes, labels, or categories that each realization of $\bm{X}$ can be assigned to. An example input space may comprise a set of raw images belonging to the categories (denoted by integers) in a given label space; 
    \item a classification model $f_{\bm{\theta}}: \mathcal{X} \to \mathcal{Y}$ with trained parameters $\bm{\theta}$. This classification model, which has been trained on $\mathcal{D_{\mathrm{train}}}$ and validated on $\mathcal{D_{\mathrm{val}}}$, produces the predicted class label $\hat{y} = \argmax_{y \in \mathcal{Y}}P(Y=y | \bm{X} = \displaystyle \vx, \bm{\theta}$) and the corresponding predicted probability $\hat{p} = \max_{y \in \mathcal{Y}}P(Y=y | \bm{X} = \displaystyle \vx, \bm{\theta}$); and
    \item a second test dataset, 
    $\mathcal{D_{\mathrm{test2}}} = \{ (\displaystyle \vx^{(i)}, y^{(i)})\}_{i=1}^{M} \subset \mathcal{X \times Y}$,
    such that $\mathcal{D} \cap \mathcal{D_{\mathrm{test2}}} = \emptyset$.
    \item an $i.i.d.$ dataset $\mathcal{D} = \{ (\displaystyle \vx^{(i)}, y^{(i)})\}_{i=1}^{N} \subset \mathcal{X \times Y}$, which is further partitioned into disjoint train ($\mathcal{D_{\mathrm{train}}}$), validation ($\mathcal{D_{\mathrm{val}}}$), and test ($\mathcal{D_{\mathrm{test1}}}$)\footnote{The ImageNet dataset does not include the ground truth labels for the test partition. Therefore, the accuracy on the validation partition is usually reported in published articles.} datasets, where $\mathcal{X}$ and $\mathcal{Y} = \{1, 2, ..., K\}$ denote the input space and label space, respectively, and where $K$ represents the number of classes, labels, or categories that each realization of $\bm{X}$ can be assigned to. An example input space may comprise a set of raw images belonging to the categories (denoted by integers) in a given label space; 
    \item a classification model $f_{\bm{\theta}}: \mathcal{X} \to \mathcal{Y}$ with trained parameters $\bm{\theta}$. This classification model, which has been trained on $\mathcal{D_{\mathrm{train}}}$ and validated on $\mathcal{D_{\mathrm{val}}}$, produces the predicted class label $\hat{y} = \argmax_{y \in \mathcal{Y}}P(Y=y | \bm{X} = \displaystyle \vx, \bm{\theta}$) and the corresponding predicted probability $\hat{p} = \max_{y \in \mathcal{Y}}P(Y=y | \bm{X} = \displaystyle \vx, \bm{\theta}$); and
    \item a second test dataset, 
    $\mathcal{D_{\mathrm{test2}}} = \{ (\displaystyle \vx^{(i)}, y^{(i)})\}_{i=1}^{M} \subset \mathcal{X \times Y}$,
    such that $\mathcal{D} \cap \mathcal{D_{\mathrm{test2}}} = \emptyset$.
\end{compactitem}

\begin{algorithm}[t]
\small
\caption{Strategy for matching two datasets based on both the predicted labels and probabilities of a model}
\label{algo:matching_strategy}
\begin{algorithmic}
    \Procedure {MatchPredictions}{$f_{\bm{\theta}}$, $\mathcal{D_{\mathrm{src}}}$, $\mathcal{D_{\mathrm{tgt}}}$, $\epsilon$}
    \State $\mathcal{P}_{\mathrm{src}} \leftarrow \Call{Predict}{f_{\bm{\theta}}, \mathcal{D_{\mathrm{src}}}}$
    \State $\mathcal{P}_{\mathrm{tgt}} \leftarrow \Call{Predict}{f_{\bm{\theta}}, \mathcal{D_{\mathrm{tgt}}}}$ 
    \vspace{10pt}
    \State Initialize 3 empty arrays:
    \State $\mathcal{P}_{\mathrm{src\_matched}}$ = [ ]
    \State $\mathcal{P}_{\mathrm{tgt\_matched}}$ = [ ]
    \State $\mathcal{P}_{\mathrm{tgt\_unmatched}}$ = [ ]
    \vspace{10pt}
    \State $N \leftarrow \Call{Length}{\mathcal{P}_{\mathrm{tgt}}}$
    \vspace{10pt}
        
    \For{$i\gets 1, N$}
        \State $y_{t}, \hat{y}_{t}, \hat{p}_{t} \leftarrow \mathcal{P}_{\mathrm{tgt}}[i]$
        \State $y_{s}, \hat{y}_{s}, \hat{p}_{s}, is\_match \leftarrow \Call{Match}{y_{t}, \hat{y}_{t}, \hat{p}_{t},\mathcal{P}_{\mathrm{src}}, \epsilon}$
        \If{$is\_match$}
            \State $\mathcal{P}_{\mathrm{src\_matched}}.append((y_{s}, \hat{y}_{s}, \hat{p}_{s}))$
            \State $\mathcal{P}_{\mathrm{tgt\_matched}}.append((y_{t}, \hat{y}_{t}, \hat{p}_{t}))$
        \Else
            \State $\mathcal{P}_{\mathrm{tgt\_unmatched}}.append((y_{t}, \hat{y}_{t}, \hat{p}_{t}))$
        \EndIf
    \EndFor
        
    \Return {$\mathcal{P}_{\mathrm{src\_matched}}, \mathcal{P}_{\mathrm{tgt\_matched}}, \mathcal{P}_{\mathrm{tgt\_unmatched}}$}
    \EndProcedure 
    \newline
        
    \vspace{10pt}
    \Function{Match}{$y_{t}, \hat{y}_{t}, \hat{p}_{t}, \mathcal{P}_{\mathrm{src}}, \epsilon$}
    \State Initialize an empty $\mathcal{P}_{\mathrm{matched}}$ = [ ] 
    \ForAll {$(y, \hat{y}, \hat{p}) \in \mathcal{P}_{\mathrm{src}}$}
        
        \If{$\hat{y}_{t} = \hat{y}$ and $\hat{p}_{t} \in [\hat{p} - \epsilon, \hat{p} + \epsilon]$}
            \State $\mathcal{P}_{\mathrm{matched}}.append((y, \hat{y}, \hat{p}))$
        \EndIf
    \EndFor
    \If{|$\mathcal{P}_{\mathrm{matched}}$| > 0}
        \State $(y_{s}, \hat{y}_{s}, \hat{p}_{s}) \gets \Call{RandomSelect}{\mathcal{P}_{\mathrm{matched}}}$
        \State $\mathcal{P}_{\mathrm{src}}.remove((y, \hat{y}, \hat{p}))$  
        \State \Return $(y_{s}, \hat{y}_{s}, \hat{p}_{s}, is\_match)$
    \Else 
        \State \Return $(0, 0, 0, False)$
    \EndIf
    \EndFunction
    \newline
    \Comment The \Call{Predict}{.} function returns the predictions of a model in an array that consists of tuples $(y, \hat{y}, \hat{p})$, where $y$, $\hat{y}$, and $\hat{p}$ denote the ground truth label, the predicted label, and the predicted probability, respectively.
    \newline 
    \Comment The \Call{RandomSelect}{.} function returns a randomly selected tuple $(y, \hat{y}, \hat{p})$ from the array that contains the input tuples.
    \newline 
    \Comment The $\mathcal{P}_{\mathrm{src}}.remove((y, \hat{y}, \hat{p}))$ in \Call{RandomSelect}{.} accounts for random selection without replacement.
    \end{algorithmic}
\end{algorithm}

The task at hand is to compare the effectiveness of $f{_{\bm{\theta}}}$ on $\mathcal{D}_{\mathrm{test1}}$ and 
$\mathcal{D_{\mathrm{test2}}}$ using certain metrics, such as accuracy.

\subsection{Matching by Predicted Labels and Probabilities}

In Section~\ref{section:introduction}, we introduced two matching criteria: ($i$) matching by both the predicted labels and their corresponding predicted probabilities and ($ii$) matching by the predicted probabilities only.
In what follows, we provide the description of our proposed evaluation protocol based on the first matching criterion, i.e., matching by both the predicted labels and their corresponding predicted probabilities. Given two test datasets $\mathcal{D_{\mathrm{test1}}}$ and $\mathcal{D_{\mathrm{test2}}}$, 
our evaluation strategy aims at assessing the accuracy of a model on these datasets, while taking into account the predicted probabilities generated by this model. For convenience, we designate the dataset with the larger number of datapoints as the source dataset $\mathcal{D_{\mathrm{src}}}$, and the other dataset as the target dataset $\mathcal{D_{\mathrm{tgt}}}$. 

\begin{figure*}[tp]
\vspace{-1.5cm}
\centering
    \includegraphics[width=0.8\textwidth]{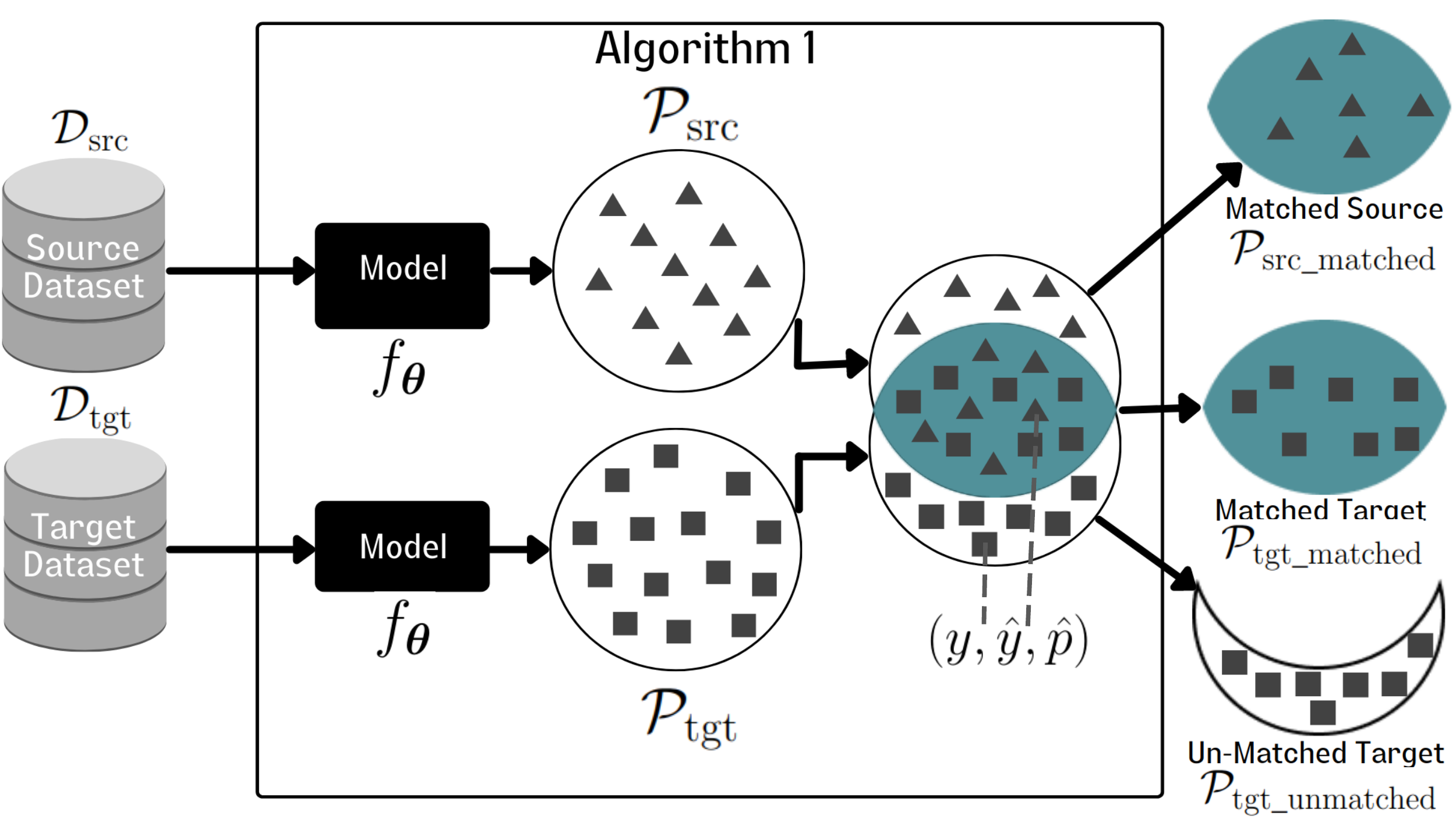}
  \caption{Visual overview of the matching strategy described by Algorithm~\ref{algo:matching_strategy}.}
\label{fig:overview_of_matching_strategy}
\end{figure*}

First, the model of interest is used to generate the predicted labels and probabilities for all the datapoints in both test datasets. We represent the output of the model, for a datapoint, as a tuple $(y, \hat{y}, \hat{p})$\footnote{Even though the $\hat{y}$ and $\hat{p}$ are sufficient for the generation of matched subsets, since we need to calculate the accuracy on the matched subsets, we adopt a tuple representation for the model output obtained for each datapoint, with this representation including the ground truth label.}, where $y$, $\hat{y}$, and $\hat{p}$ represent the ground truth label, the predicted label, and the predicted probability, respectively. 
In a next step, we generate the subsets $\mathcal{P_{\mathrm{src}}}$ and $\mathcal{P_{\mathrm{tgt}}}$, whose elements are tuples $(y, \hat{y}, \hat{p})$, by sub-sampling from all the model outputs obtained for the datasets $\mathcal{D_{\mathrm{src}}}$ and $\mathcal{D_{\mathrm{tgt}}}$, making use of predefined matching criteria. Our first matching criterion is summarized as follows: for every element in $\mathcal{P_{\mathrm{tgt}}}$, find an element in $\mathcal{P_{\mathrm{src}}}$, such that they both have the same predicted labels and such that their predicted probabilities are approximately equal. In this context, we make use of a very small fraction $\epsilon$ to control how approximately matched the probability values should be. For example, if we want the probability difference to be within $\pm1\%$, we set $\epsilon=0.01$. Our evaluation protocol, using this matching criterion, is described in more detail in Algorithm~\ref{algo:matching_strategy} and a summarizing overview is presented in Figure~\ref{fig:overview_of_matching_strategy}. The description of our evaluation protocol, using the second matching criterion -- matching by predicted probabilities only -- is described in Algorithm~\ref{algorithm_second_matching_strategy} in Appendix~\ref{appendix_description_second_matching_criterion}. 

\subsection{Accuracy Evaluation}

The outputs generated by Algorithm~\ref{algo:matching_strategy} and Algorithm~\ref{algorithm_second_matching_strategy} are arrays: $\mathcal{P}_{\mathrm{src\_matched}},$ $\mathcal{P}_{\mathrm{tgt\_matched}}$, and $\mathcal{P}_{\mathrm{tgt\_unmatched}}$. We could also generate the source datapoints that remain unmatched. However, since the source dataset is mostly the $i.i.d.$ test dataset in our experiments and since our goal is to investigate the accuracy drop in the target dataset (the replicate dataset), we excluded the unmatched source datapoints from further analysis. The $\mathcal{P}_{\mathrm{src\_matched}}$ and $\mathcal{P}_{\mathrm{tgt\_matched}}$ arrays contain information about the model outputs that meet the matching criteria, while the $\mathcal{P}_{\mathrm{tgt\_unmatched}}$ array contains information about the model outputs that do not meet the matching criteria (i.e., for the datapoints in $\mathcal{D_{\mathrm{tgt}}}$ that do not have corresponding matching datapoints in $\mathcal{D_{\mathrm{src}}}$).

Although many other metrics can be used to evaluate the aforementioned arrays to get a better understanding of the datapoints that are considered similar or not similar, as judged by the outputs of a model, the scope of this paper is restricted to the accuracy metric, given that this metric is widely used for evaluating models on multiple test datasets. Our definition of accuracy is conventional -- the ratio of the number of correctly classified datapoints to the total number of datapoints under evaluation. However, we make a distinction between accuracy evaluated on all the datapoints in a source or target dataset, and the accuracy evaluated on the matched datapoints. We refer to the former as simply {\it accuracy} and to the latter as {\it matched accuracy}.

\section{Related Research}
\label{section:related_research}

In this section, we briefly summarize research efforts that are related to our work.

\paragraph{Possible Explanations for the Accuracy Degradation in Replication Datasets.}
Even after carefully following the way the original datasets were constructed, the creators of the CIFAR-10.1 and the ImageNetV2 datasets~\cite{Recht_Do_ImageNet} concluded that the degradation in accuracy is not resulting from adaptive over-fitting to the original test datasets, suggesting that a possible explanation for the degradation is the presence of harder-to-classify images in the newer test datasets. \cite{Engstrom_Identifying_Statistical_2020} concluded that the accuracy degradation in the ImageNetV2 dataset is the result of statistical bias in the creation of the ImageNetV2 dataset. They show that the accuracy gap is indeed narrower if various sources of statistical bias are accounted for. Specifically, they demonstrate that only $3.6\%\pm 1.5\%$ of the accuracy drop for ImageNetV2 remains unaccounted for. Our experimental results support their finding and our proposed evaluation protocol additionally provides a practical tool for realistic performance assessment of DNN models on multiple test datasets.

\paragraph{Softmax as Baseline for DNN Model Uncertainty.}
There are studies that consider softmax outputs as approximations to the probabilities that are generated by DNN classification models: the use of softmax outputs has been considered as a baseline for detecting misclassified and out-of-distribution datapoints by~\cite{hendrycks_a_baseline_for_2019}, and has been demonstrated to be decent at capturing the uncertainty in the predictions produced by DNN models~\cite{mukhoti_calibrating_deep_2020, pearce_understanding_softmax_confidence_2021}. Our principled evaluation approach and our empirical evaluation results lend support to the use of softmax outputs as a baseline for uncertainty estimation. However, we also align with other works~\cite{Guo_on_calibration_of_2017, hein_why_relu_networks_2019} that call for caution in the interpretation of softmax outputs as confidence values or probabilities that models have about their predictions, since models are prone to over-confidence. Although well-calibrated probabilities are desirable, our approach is not explicitly aimed at evaluating how calibrated the probabilities are. We aim at investigating how consistent the outputs of a model are. For example, a model that is over-confident but also consistent in multiple environments could still be useful for downstream pipelines, and the probabilities generated by the model could be further calibrated.

\paragraph{Calibration of Models.}
Model calibration or confidence calibration is the problem of predicting probability estimates that are representative of the true correctness likelihood~\cite{Guo_on_calibration_of_2017}. Published methods for model calibration involve improving the modeling process~\cite{MMCE_Loss, maddox_a_simple_baseline_2019, mukhoti_calibrating_deep_2020, mukhoti_deterministic_neural_network_2021, ranganath_improving_model_calibration_2020}, thereby creating models that inherently produce calibrated probabilities, or utilizing post-processing techniques to improve the outputs of already trained models~\cite{naeini_obtaining_well_calibrated_2015, Guo_on_calibration_of_2017}. The Expected Calibration Error~\cite{naeini_obtaining_well_calibrated_2015, Guo_on_calibration_of_2017} is a popular metric for evaluating how calibrated models are. Although popular, the ECE has a number of weaknesses: ($i$) given that ECE is histogram-based, it heavily depends on the binning strategy used~\cite{nixon_measuring_calibration_in_2019, ding_evaluation_of_neural_2019}; ($ii$) the predicted probabilities generated by different models for the same dataset could have different densities, hence making the ECE not the ideal metric for comparing the calibration of a model on different datasets; and ($iii$) ECE is independent of accuracy, which means that a model could be calibrated but have low accuracy, and vice versa, which implies the existence of a trade-off between accuracy and calibration. Our evaluation approach jointly evaluates the accuracy and consistency of the predicted probabilities in relation to accuracy, thereby enabling an effective evaluation of a model on multiple similar-but-non-identical datasets.

\paragraph{Classification with a Reject Option.}
Classifiers with a reject option can abstain from predicting a label if certain criteria are not met. Examples of classifiers with a reject option are those where the rejection is based on confidence thresholds~\cite{chu_optimal_decision_functions_1965, cortes_learning_with_rejection_2016}. This group of classifiers comes with a trade-off between error rates and rejection rates, benefiting from reliable confidence estimation. Whereas research efforts on classification with a reject option aim at creating classifiers that have improved effectiveness while being able to abstain from making certain predictions, the aim of our work is to evaluate any DNN classification model in a principled manner. This is even more crucial when evaluating a model under different settings.

\section{Experimental Setup and Results}
\label{section:experimental_setup_and_results}

\subsection{Experimental Setup}

\paragraph{Dataset Pairs}: The goal of our experiments is to evaluate a representative sample of published DNN classification models on the CIFAR-10 and ImageNet datasets, using the evaluation protocol proposed in Section~\ref{section:proposed_evaluation_protocol}. To that end, we trained $278$ models on the CIFAR-10 dataset using PyTorch implementations by \cite{mukhoti_calibrating_deep_2020}\footnote{We additionally used models implemented in \href{https://github.com/hysts/pytorch_image_classification]}{https://github.com/hysts/pytorch\_image\_classification}.} under various training settings. These training settings include different model architectures, loss functions, learning rate schedules, data augmentation techniques, and random seeds. We describe the datasets used in Table~\ref{table:test_dataset_description} in Appendix~\ref{appendix_description_of_datasets} and we list the various characteristics of the trained models in Table~\ref{table:models_xtics_cifar10} in Appendix~\ref{appendix_description_of_the_trained_models}. For CIFAR-10, evaluations are carried out on the following test dataset pairs: ($i$) CIFAR-10 versus CIFAR-10.1, ($ii$) CIFAR-10 versus CIFAR-10.2, and ($iii$) CIFAR-10 versus CINIC-10. The CIFAR-10.2 dataset was created by \cite{Lu_Harder_or} and is a variant of the CIFAR-10 dataset; it was derived from the same source as CIFAR-10 and assembled via a similar process. The CINIC-10 dataset\cite{Darlow_Cinic_10} was derived from ImageNet, and consists of images that belong to the same classes as those of the CIFAR-10 dataset; all the images where resized to the resolution of $32$$\times$$32$.

Due to computational limitations, we did not re-train the ImageNet models but utilized $286$ pre-trained published models from a popular PyTorch repository~\cite{Timm_Models_Repo}. The network architectures of the pre-trained ImageNet models and their names, as used in the source GitHub repository, are presented in Table~\ref{table:models_xtics_imagenet} in Appendix~\ref{appendix_description_of_the_trained_models}. Similar to what was done for the CIFAR-10 pairs, we determine the accuracy and matched accuracy for the ImageNetV1 versus ImageNetV2 dataset pair for each pre-trained model. The ImageNetV1 dataset consists of a total of $50,000$ images taken from $1,000$ categories, while ImageNetV2 consists of a total of $10,000$ images taken from $1000$ categories.



\begin{figure}[!htb]
\subfloat[\label{fig:acc_matched_acc_cifar10_vs_cifar10.1_1st_match_criterion}]{%
  \includegraphics[ width=0.45\linewidth]{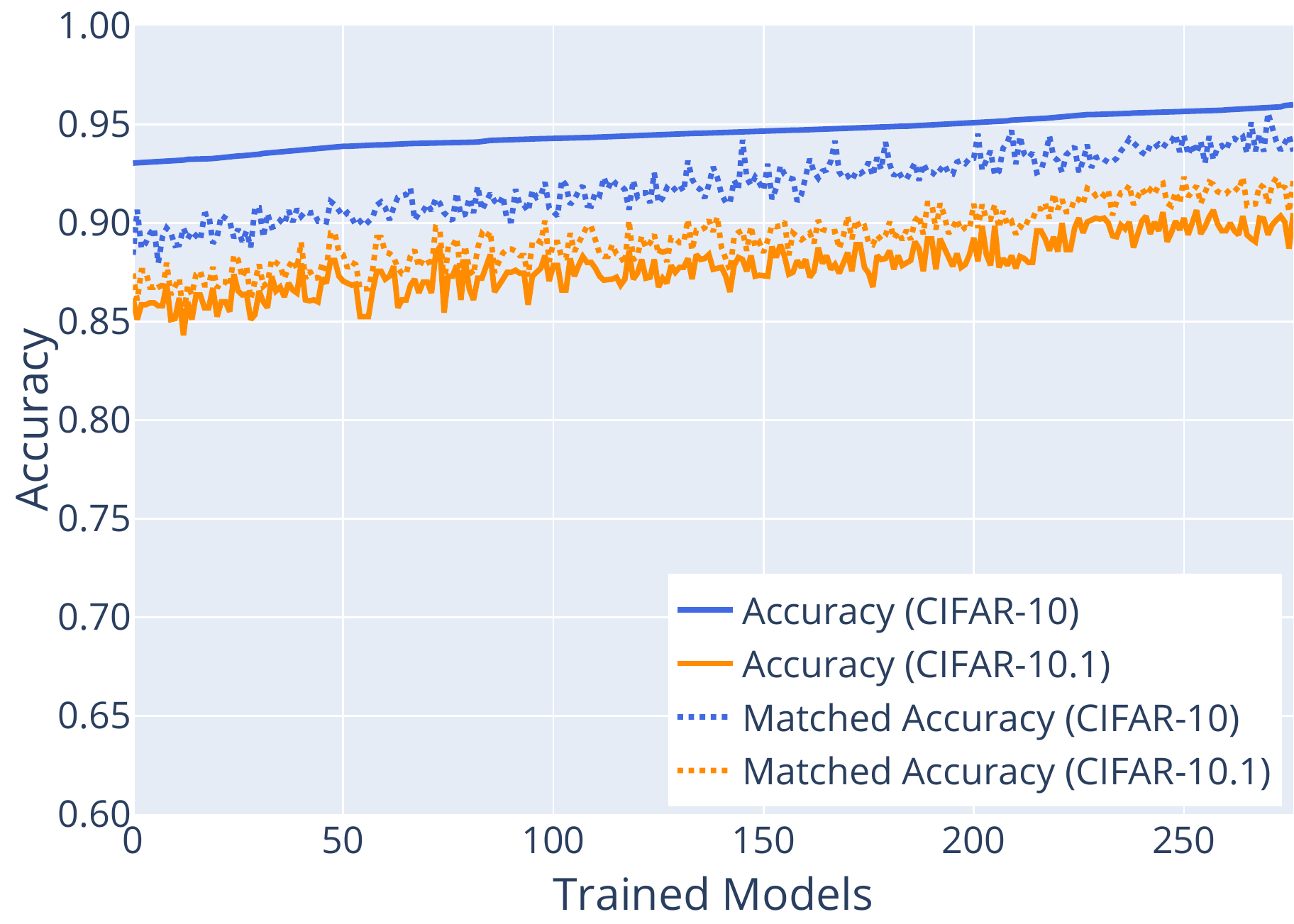}}\hfill
\subfloat[\label{fig:xtics_cifar10_vs_cifar10.1_1st_match_criterion}]{%
  \includegraphics[ width=0.45\linewidth]{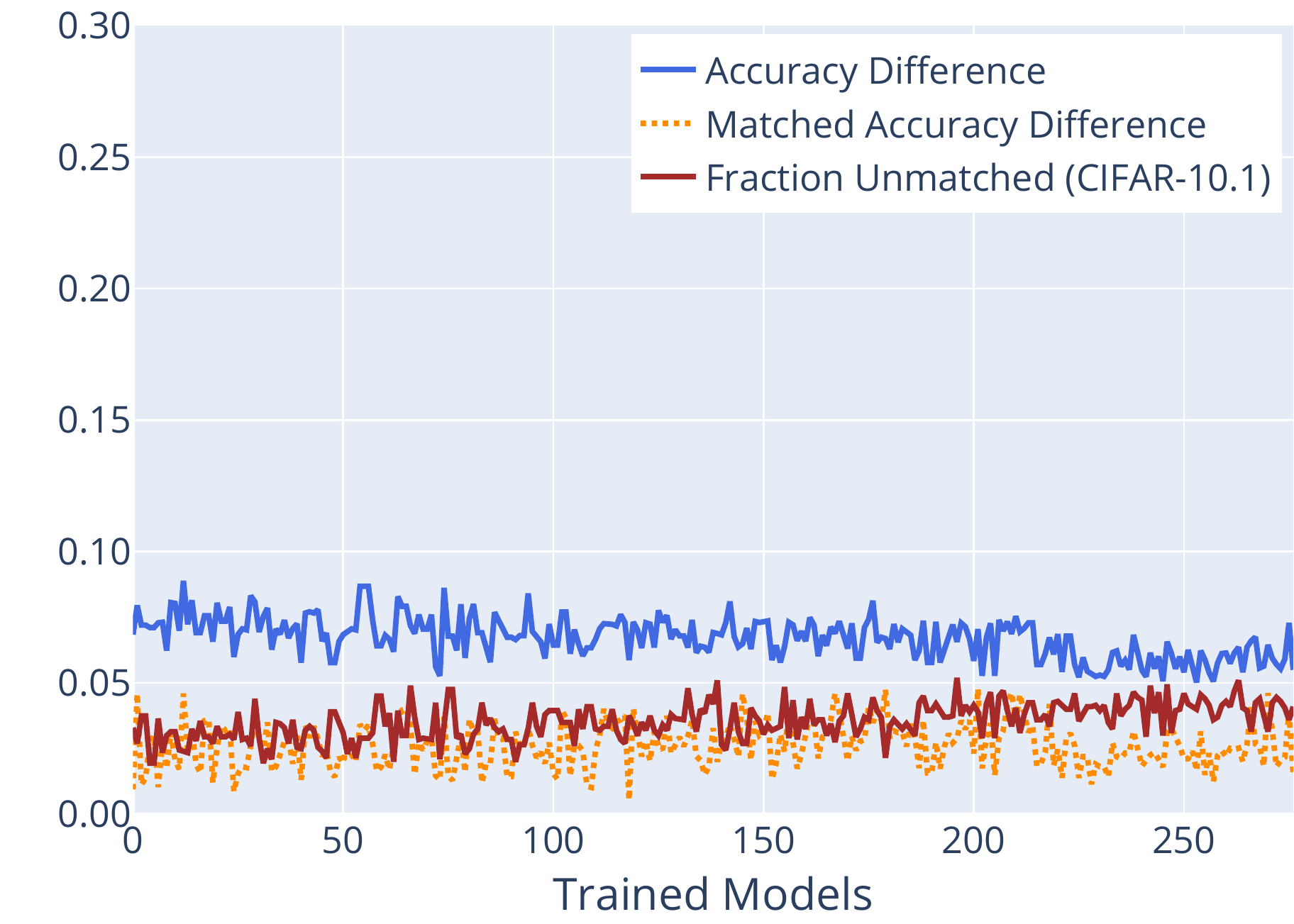}}\vfill
\subfloat[\label{fig:acc_matched_acc_cifar10_vs_cifar10.2_1st_match_criterion}]{%
  \includegraphics[ width=0.45\linewidth]{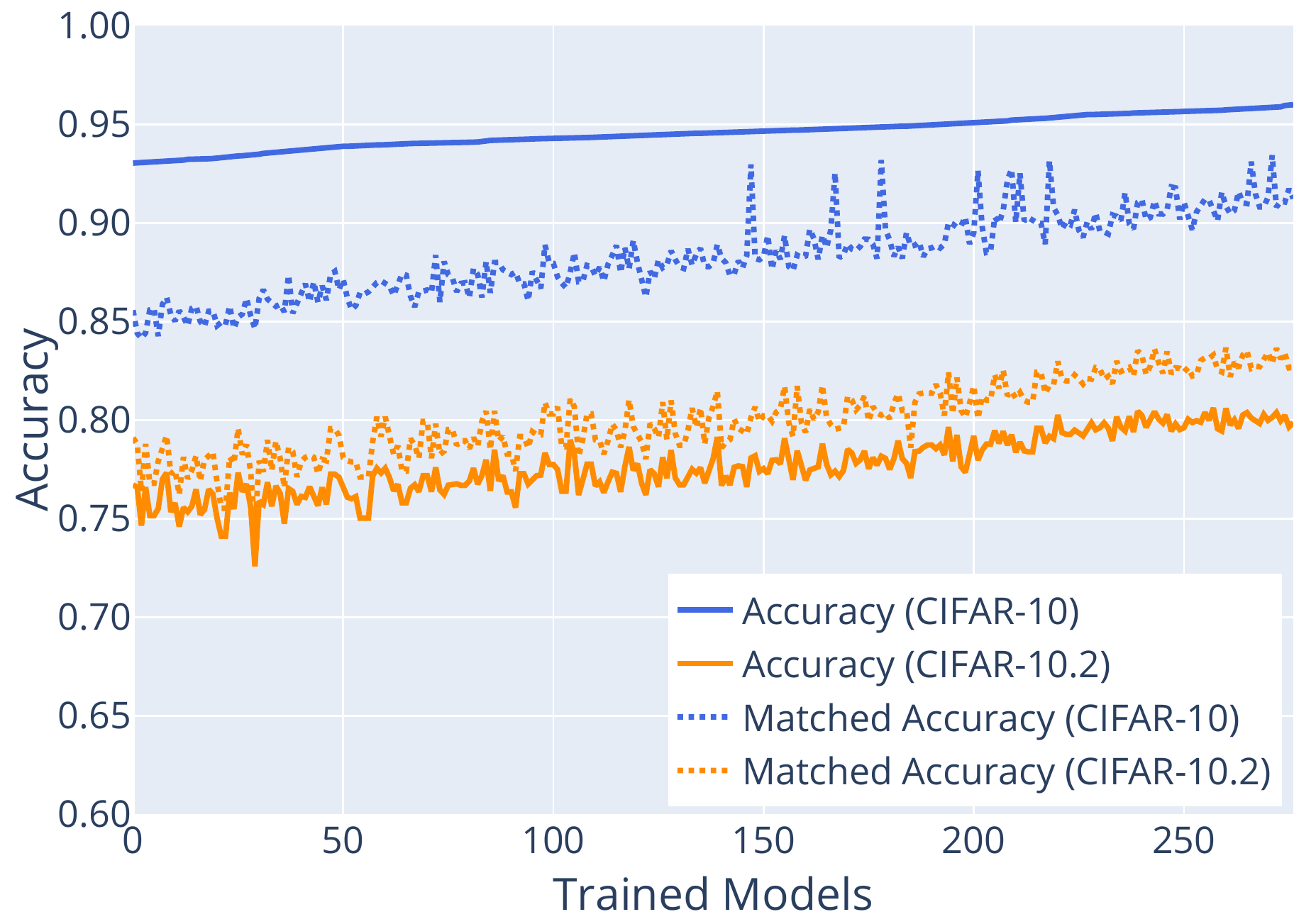}}\hfill
\subfloat[\label{fig:xtics_cifar10_vs_cifar10.2_1st_match_criterion}]{%
  \includegraphics[ width=0.45\linewidth]{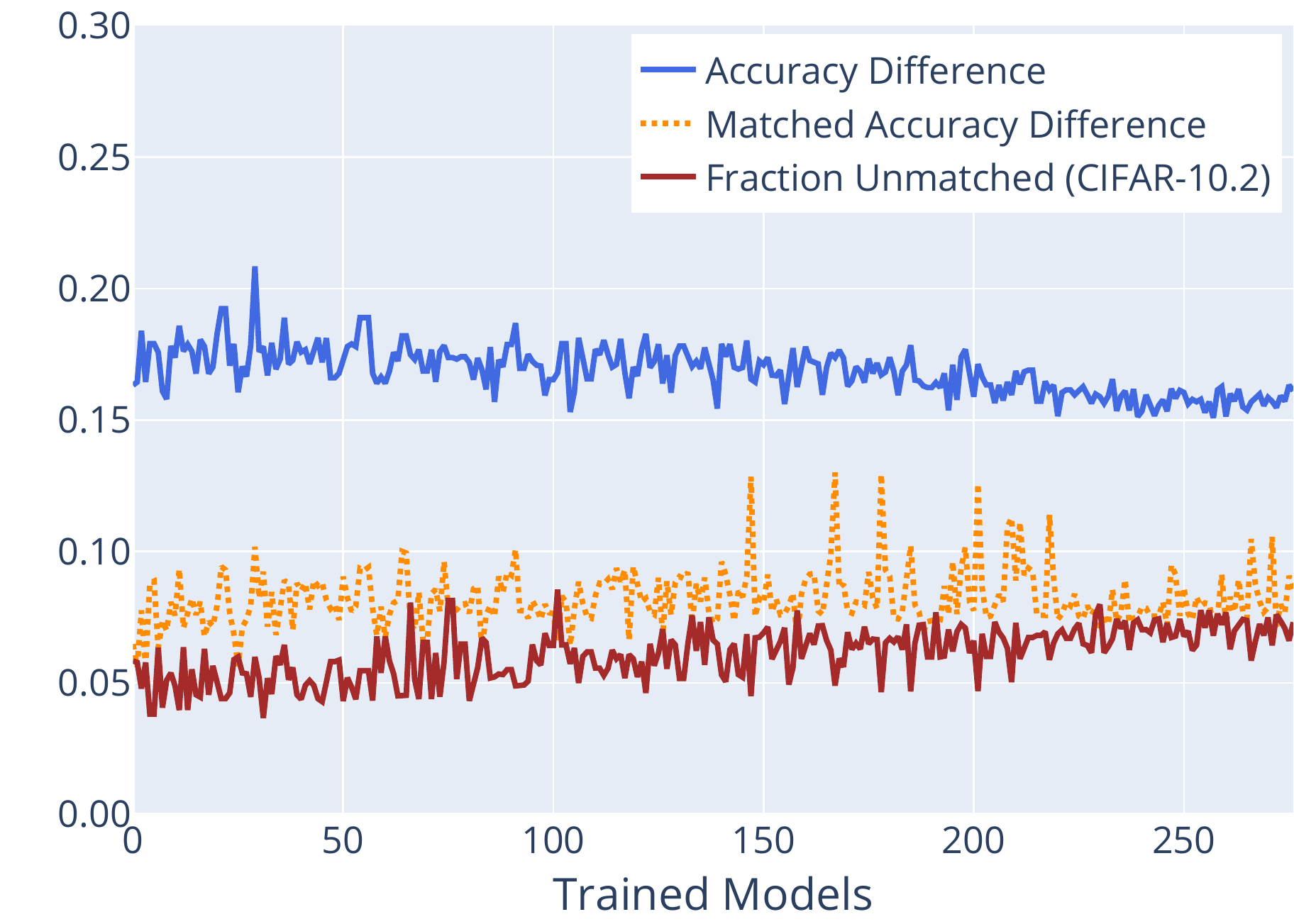}}\vfill
  \subfloat[\label{fig:acc_matched_acc_cifar10_vs_cinic10_1st_match_criterion}]{%
  \includegraphics[ width=0.45\linewidth]{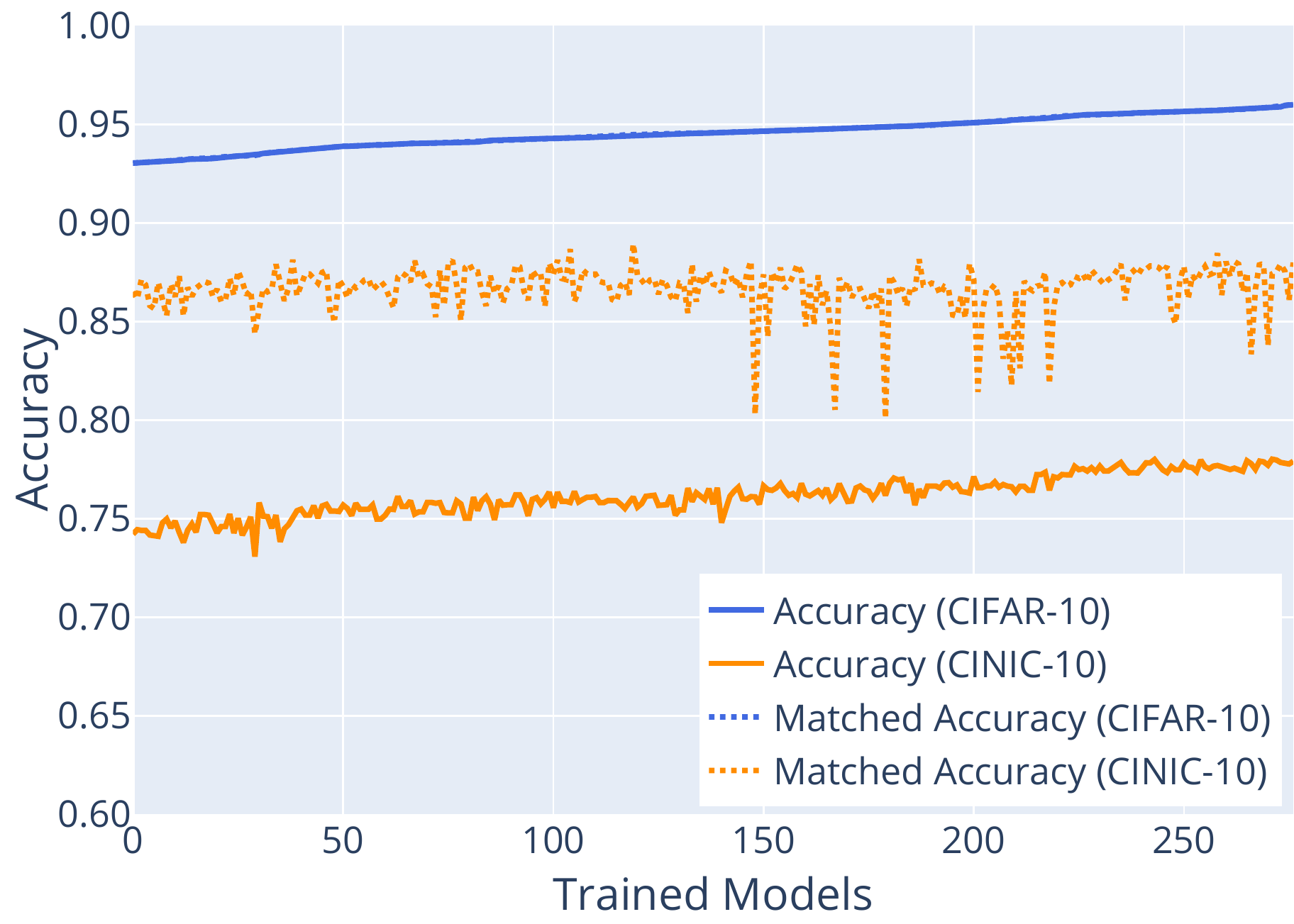}}\hfill
\subfloat[\label{fig:xtics_cifar10_vs_cinic10_1st_match_criterion}]{%
  \includegraphics[ width=0.45\linewidth]{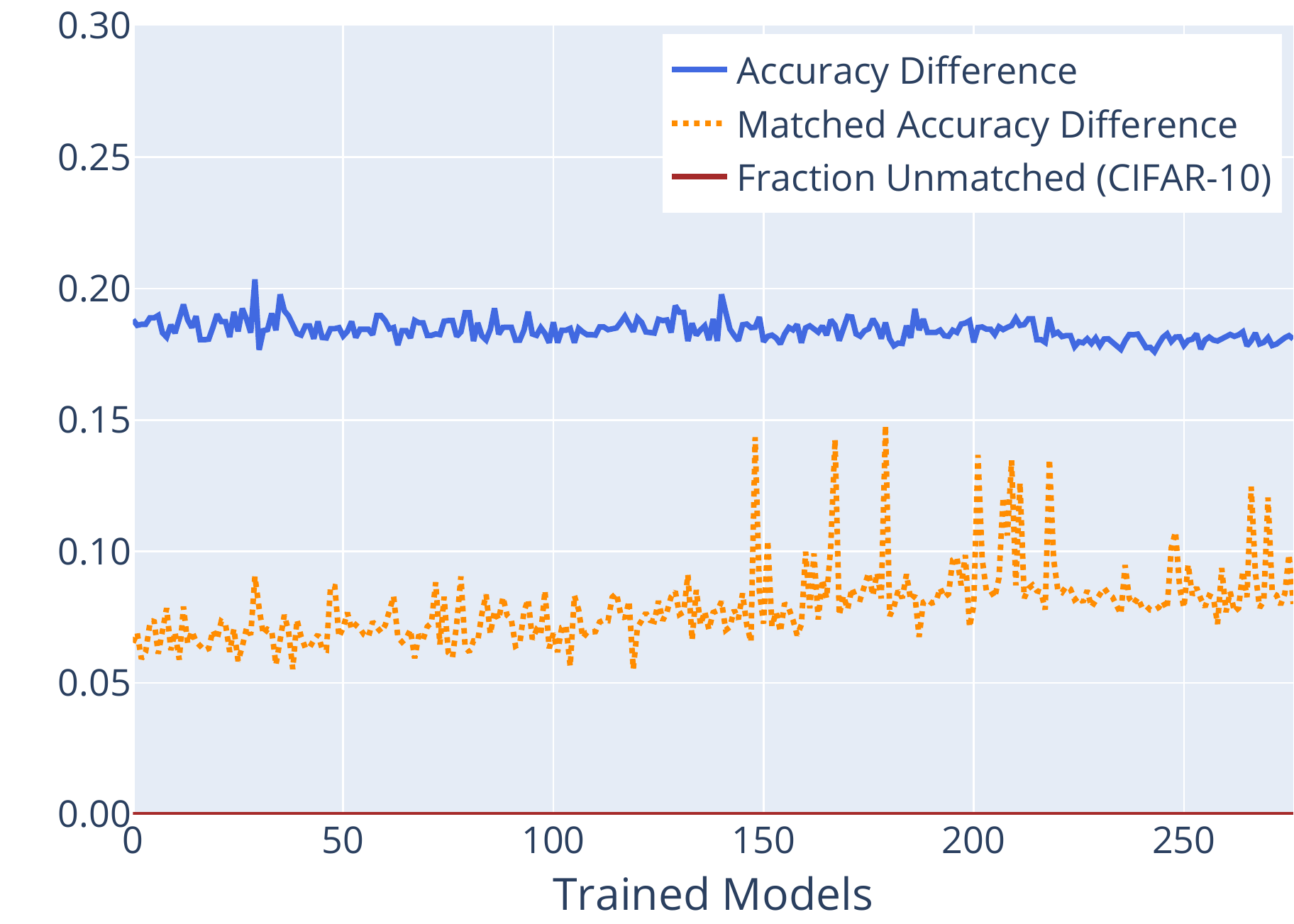}}\vfill
\caption{Plots for the accuracy and matched accuracy for (a) CIFAR-10 versus CIFAR-10.1, (c) CIFAR-10 versus CIFAR-10.2, and (e) CIFAR-10 versus CINIC-10. Plots for accuracy difference, matched accuracy difference, and fraction unmatched for (b) CIFAR-10 versus CIFAR-10.1, (d) CIFAR-10 versus CIFAR-10.2, and (f) CIFAR-10 versus CINIC-10. The models in all plots are sorted according to increasing accuracy on CIFAR-10.}
\label{fig:acc_matched_acc_xtics_cifar_related_1st_match_criterion}
\vspace{-10pt}
\end{figure}

\begin{figure}[!htb]
\subfloat[\label{fig:acc_matched_acc_imagenet_vs_imagenetv2_1st_match_criterion}]{%
  \includegraphics[width=0.48\linewidth]{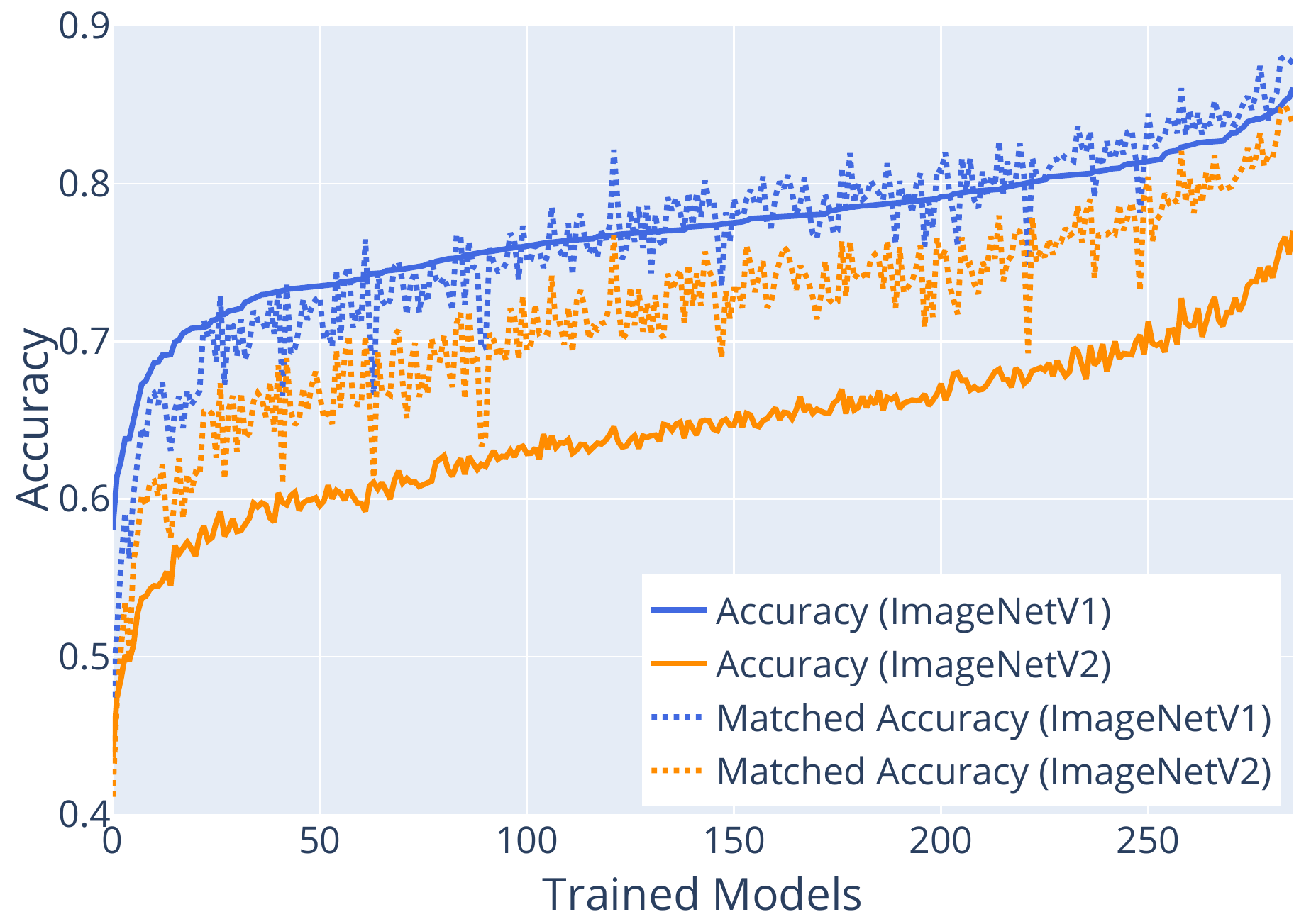}}\hfill
\subfloat[\label{fig:xtics_imagenet_vs_imagenetv2_1st_match_criterion}]{%
  \includegraphics[width=0.48\linewidth]{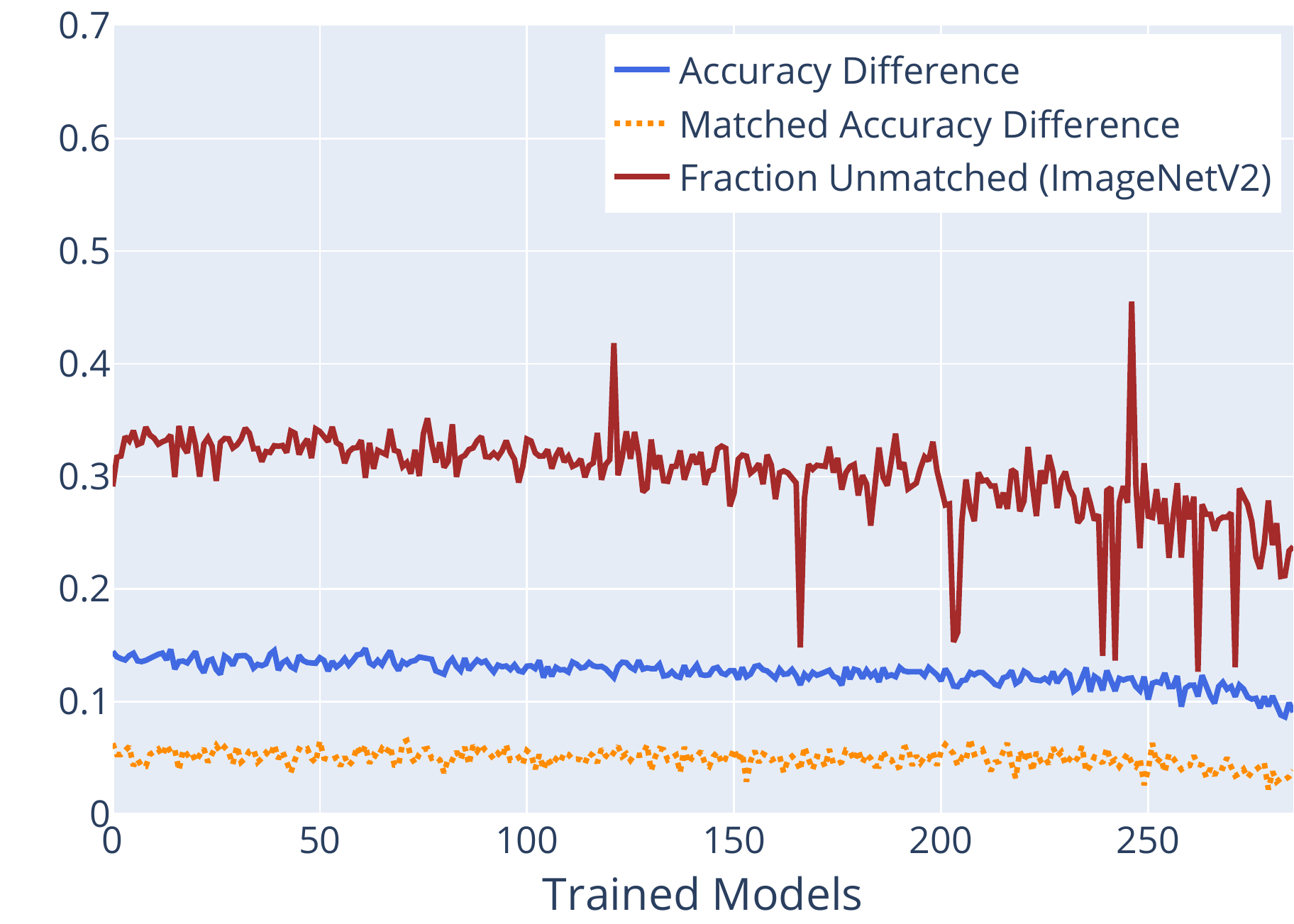}}

\caption{(a) Accuracy and matched accuracy plots and (b) difference in accuracy and matched accuracy for $286$ models evaluated on the ImageNet and ImageNetV2 dataset pair according to Algorithm~\ref{algo:matching_strategy}. The models in all plots are sorted according to increasing accuracy on ImageNetV1.}
    \label{fig:acc_matched_acc_xtics_imagenet_related_1st_match_criterion}
\end{figure}

\paragraph{Generated Results.} We applied Algorithm~\ref{algo:matching_strategy} and Algorithm~\ref{algorithm_second_matching_strategy} to each of the aforementioned dataset pairs to generate matching subsets, with $\epsilon = 0.005$. Subsequently, each model is evaluated on these subsets to determine ($i$) the accuracy on the source and the target datasets and ($ii$) the matched accuracy on the matched source and matched target datasets. Another value calculated is the fraction of datapoints in the target dataset that do not find a match with datapoints in the source dataset, which we refer to as {\it Fraction Unmatched}. These results are presented in Figure~\ref{fig:acc_matched_acc_xtics_cifar_related_1st_match_criterion} and Figure~\ref{fig:acc_matched_acc_xtics_imagenet_related_1st_match_criterion}. Additional results can be found in Figure~\ref{fig:acc_matched_acc_xtics_cifar_related_2nd_match_criterion} and Figure~\ref{fig:acc_matched_acc_xtics_imagenet_related_2nd_match_criterion} in Appendix~\ref{appendix_results_from_matching_by_only_the_predicted_probabilities}. The accuracy and matched accuracy gaps are better visualized in the plots on the right column of these Figures.
The results for the matched accuracy are the average of 10 runs of each algorithm (Algorithm~\ref{algo:matching_strategy} and Algorithm~\ref{algorithm_second_matching_strategy}). Since the standard errors over these runs were negligible, we did not include error bars for the matched accuracy. For accuracy, we evaluate already trained models and it is not feasible to have the same pre-trained models over multiple seeds. Therefore, we plot the accuracy of the available pre-trained models.

\subsection{Results and Discussion}

\paragraph{Consistent and Significant Accuracy Drops.} Our extensive evaluation of the accuracy of a plethora of models, on both the original and replication datasets, produced results that are in line with the results reported by~\cite{Recht_Do_ImageNet}, \cite{John_Accuracy_on}, and \cite{Lu_Harder_or}. When accuracy is evaluated on all datapoints, we observe consistent but significant accuracy drops on the replicated datasets. 
These results are shown in Figures~\ref{fig:acc_matched_acc_cifar10_vs_cifar10.1_1st_match_criterion}, \ref{fig:acc_matched_acc_cifar10_vs_cifar10.2_1st_match_criterion}, \ref{fig:acc_matched_acc_cifar10_vs_cinic10_1st_match_criterion}, and \ref{fig:acc_matched_acc_imagenet_vs_imagenetv2_1st_match_criterion} as the gap between the solid blue line and the solid orange line; the difference in accuracy can be better viewed in the corresponding accuracy difference plots in Figures~\ref{fig:xtics_cifar10_vs_cifar10.1_1st_match_criterion}, \ref{fig:xtics_cifar10_vs_cifar10.2_1st_match_criterion}, \ref{fig:xtics_cifar10_vs_cinic10_1st_match_criterion}, and \ref{fig:xtics_imagenet_vs_imagenetv2_1st_match_criterion} as the blue line.




\paragraph{Narrower Matched Accuracy Gaps.}
Given an input image, a model is used to predict the label for this image and the likelihood of the correctness of the prediction, which we loosely interpret as the belief of the model about the correctness of the prediction. Algorithm~\ref{algo:matching_strategy} enables us to effectively partition the evaluation datasets into matching and non-matching subsets and to evaluate the consistency of the accuracy of models with respect to uncertainty. 

By utilizing our evaluation protocol, which allows taking into account uncertainty information, the matched accuracy gap is consistently narrower, indicating that the accuracy of the models on the test dataset pairs are closer if the uncertainty information is taken into account. The key takeaway message here is that the uncertainty that models generates provide additional information that should be taken into to better evaluate the models. These results are presented in Figures~\ref{fig:acc_matched_acc_cifar10_vs_cifar10.1_1st_match_criterion}, \ref{fig:acc_matched_acc_cifar10_vs_cifar10.2_1st_match_criterion}, \ref{fig:acc_matched_acc_cifar10_vs_cinic10_1st_match_criterion}, and \ref{fig:acc_matched_acc_imagenet_vs_imagenetv2_1st_match_criterion} as the gap between the dotted blue line and the dotted orange line; the accuracy difference can better be viewed in Figures~ \ref{fig:xtics_cifar10_vs_cifar10.1_1st_match_criterion}, \ref{fig:xtics_cifar10_vs_cifar10.2_1st_match_criterion}, \ref{fig:xtics_cifar10_vs_cinic10_1st_match_criterion}, and \ref{fig:xtics_imagenet_vs_imagenetv2_1st_match_criterion} as the orange line. The evaluation results obtained when matching is done by predicted probabilities only (our second matching criterion) can be found in Figures~\ref{fig:acc_matched_acc_xtics_cifar_related_2nd_match_criterion} and \ref{fig:acc_matched_acc_xtics_imagenet_related_2nd_match_criterion} in Appendix~\ref{appendix_description_second_matching_criterion}. 

\begin{figure}[!htb]
\subfloat[\label{fig:ig_resnext101_32x48dcal_curves}]{%
  \includegraphics[width=0.47\linewidth]{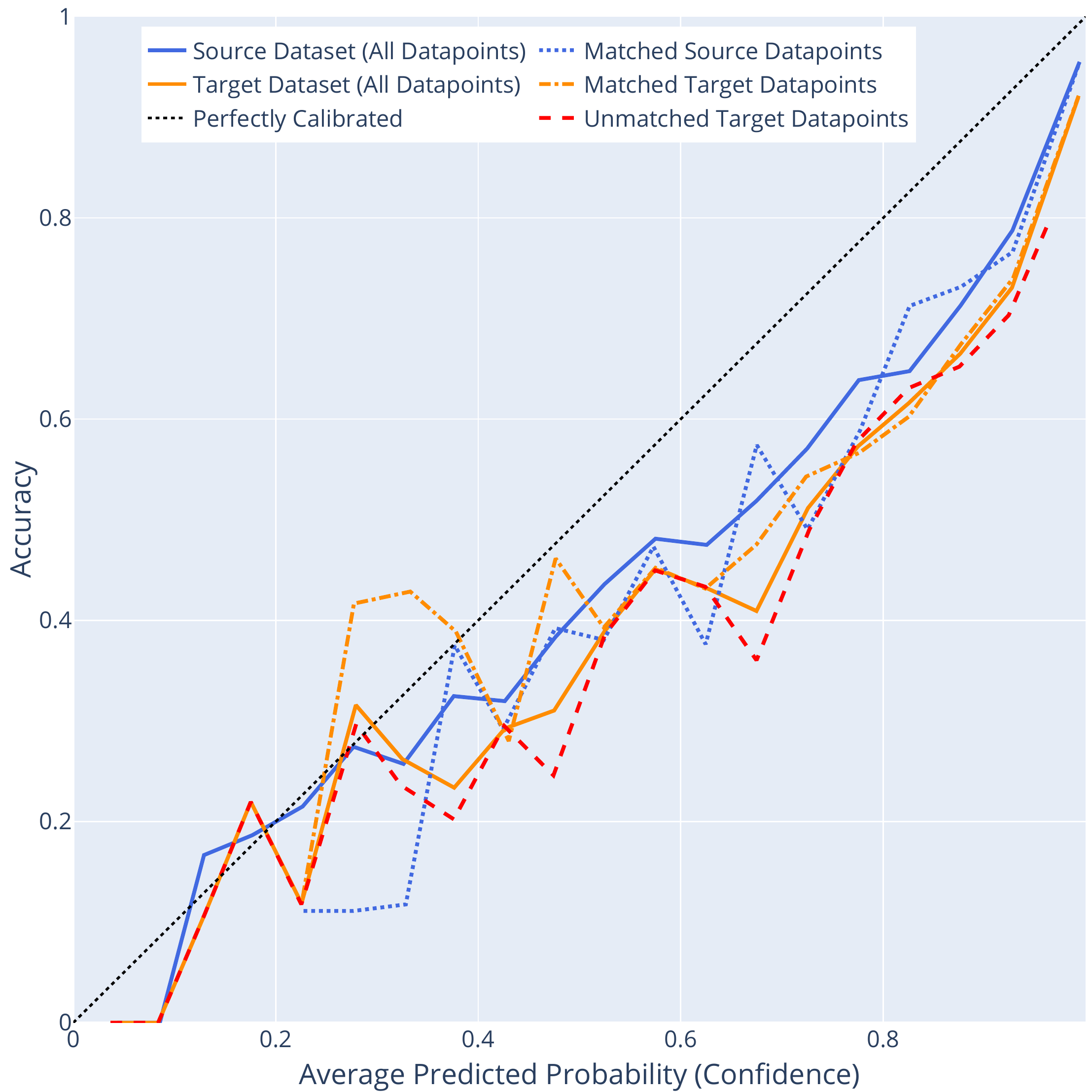}}\hfill
\subfloat[\label{fig:ig_resnext101_32x48d_densities}]{%
  \includegraphics[width=0.51\linewidth]{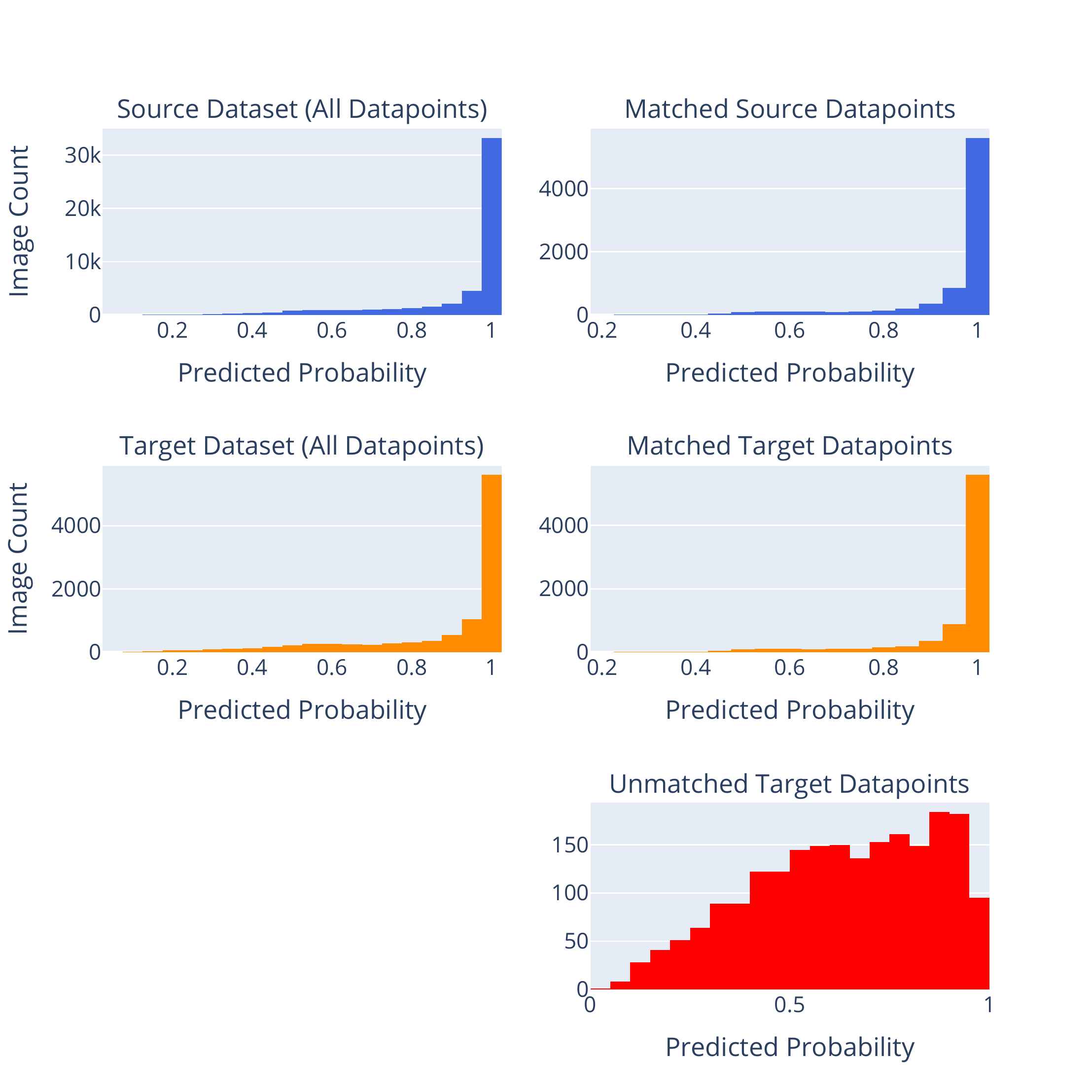}}

\caption{The plots of characteristics of a pre-trained ResNext101 model~\cite{Mahajan_Exploring_the_limits_2018} when evaluated on various subsets of the ImageNetV1 and ImageNetV2 datasets, as generated by our proposed evaluation protocol: (a) accuracy versus confidence and (b) density plots for the predicted probabilities.}

    \label{fig:cal_curves_and_density_plots_ig_resnext101_32x48d}
\end{figure}

\paragraph{Consistency of Confidence versus Accuracy across Various Subsets of the Test Datasets.} To further illustrate the behavior of models for the various dataset subsets, we present in Figure~\ref{fig:ig_resnext101_32x48dcal_curves} the accuracy versus confidence curves, which are also known as calibration curves or reliability diagrams~\cite{Mizil_Predicting_Good_Probabilities_2005}, for all the subsets of the ImageNetV1 and the ImageNetV2 datasets. The predictions were generated by utilizing our proposed protocol to evaluate a pre-trained ResNext101~\cite{Mahajan_Exploring_the_limits_2018} ImageNet model.\footnote{The pre-trained model, named as ig\_resnext101\_32x48d, was obtained from \href{https://github.com/rwightman/pytorch-image-models}{https://github.com/rwightman/pytorch-image-models}; this is the implementation of the model proposed in~\cite{Mahajan_Exploring_the_limits_2018}.} 
In Figure~\ref{fig:ig_resnext101_32x48d_densities}, we show the density plots for the predicted probabilities. The accuracy gap and the matched accuracy gap are $8.6\%$ and $3.1\%$, respectively, while the accuracy on the unmatched datapoints is $45.54\%$. These results primarily demonstrate that the accuracy obtained for different test sets can vary widely if some of their characteristics differ. However, the model behavior within these subsets is fairly consistent with respect to the correlation between the uncertainty values it generates and the accuracy for its predictions. This has practical implications: the performance of a deployed model can be inferred from the uncertainty values it generates, which can benefit downstream applications that rely on the output from the model. Additional plots for the accuracy versus confidence, for a semi-weakly supervised pretrained ResNet-18~\cite{ResNet, Yalniz_Billion_Scale_Semi, Timm_Models_Repo} model, and the density plots are provided in Figure~\ref{fig:_swsl_resnet18cal_curves} of Appendix~\ref{appendix_additional_examples_cal_curves_and_density_plots}.

\section{Conclusions and Future Work}
\label{section:conclusion_and_future_work}

DNN classification models are complex and are equally trained on complex data. By using simplistic evaluation protocols that do not fully account for the rich information generated by these classification models, incomplete or incorrect conclusions about their performance could be drawn. This is even more crucial when comparing the performance of models on multiple similar-but-non-identical datasets. For such evaluation scenarios, using accuracy alone to directly compare the performance of a model over multiple datasets (on all the datapoints available) might not be adequate since these datasets could have valid differences. If the predictions of a model on multiple datasets are consistent, it is fair to expect the performance of the model on matched subsets of these datasets to be similar. We believe that the consistent behaviour of a model in different environments is a desirable property.

In this paper, we introduced a principled evaluation protocol that is suitable for the comparison of the performance of models on multiple similar-but-non-identical test datasets. Specifically, by leveraging the proposed evaluation protocol to generate accuracy values for matched subsets from pairs of test datasets, we were able to perform an in-depth investigation of $278$ CIFAR-10 and $286$ ImageNet models on the original ImageNet and CIFAR-10 datasets, as well as their replication datasets. Our experimental results demonstrate that the deterioration in accuracy may not be as large as was reported by prior research efforts, if the uncertainty-related information generated by DNN models is utilized. While it is encouraging that softmax probabilities contain uncertainty-related information and that these probabilities can serve as a baseline for the uncertainty of a model, it may be valuable to further utilize the proposed evaluation protocol together with alternative uncertainty estimation techniques. Furthermore, it would be of interest to evaluate the consistency of the accuracy metric with respect to the uncertainty estimates in settings with identifiable covariate shift. Finally, we hope that the results presented in this paper will lend support to other research efforts that emphasize the importance of leveraging model uncertainty for evaluation purposes.

\newpage
\medskip

{

\bibliographystyle{unsrt}  
\bibliography{references}

}

\newpage
\appendix

\section{Appendix}

\subsection{Description of the Datasets}
\label{appendix_description_of_datasets}
\begin{table}[ht]
 \small
  \caption{Description of the dataset partitions for CIFAR-10 and its replication datasets.}
  \label{table:test_dataset_description}
  \centering
  \begin{tabular}{llll}
    \toprule
    Name  & Default Available & Partition & Total \\
          & Partitions & Used & Images \\
    \midrule
    CIFAR-10 & Train, Test  & Test   & $10,000$  \\
    CIFAR-10.1     & Test & Test   & $2,000$    \\
    CIFAR-10.2     & Test  & Test  & $10,000$ \\
    CINIC-10     & Train, Validation, Test & Test  & $90,000$ \\
    \bottomrule
  \end{tabular}
\end{table}
 
\subsection{Description of the Trained Models}
\label{appendix_description_of_the_trained_models}

\begin{table}[ht]
\small
\centering
\caption{Description of the characteristics of the trained models for CIFAR-10.}
\label{table:models_xtics_cifar10}
\begin{tabular}{p{1.7cm}l}
    \toprule
    \begin{tabular}{c}
         Network \\
         Architectures 
    \end{tabular} &
    \begin{tabular}{l}
        VGG~\cite{VGG}, ResNet~\cite{ResNet}, \\ 
        DenseNet~\cite{DenseNet},  ResNeXt~\cite{ResNeXt}, \\
        ResNet\_Preact~\cite{ResNet_Preact}, Shake\_Shake~\cite{Shake_Shake} \\
        PyramidNet~\cite{PyramidNet}, SENet~\cite{SENet} 
    \end{tabular} \\ 
    
    \midrule
    
    \begin{tabular}{c}
         Loss \\
         Functions 
    \end{tabular} &
    \begin{tabular}{l}
        Focal Loss~\cite{Focal_Loss}, Brier Loss~\cite{Brier_Loss},     \\
        Maximum Mean Calibration Error (MMCE) Loss~\cite{MMCE_Loss},     \\
        Label Smoothing Loss~\cite{Label_Smoothing_Loss} \\
    \end{tabular}   \\
    
    \midrule
    
    \begin{tabular}{c}
         Data \\
         Augmentations  \\
         (Basic)
    \end{tabular} &
    \begin{tabular}{l}
        Random Cropping,    \\
        Horizontal Flipping     \\
    \end{tabular}   \\
    \midrule
    
    \begin{tabular}{c}
         Data \\
         Augmentations  \\
         (Advanced)
    \end{tabular} &
    \begin{tabular}{l}
        CutMix~\cite{CutMix_Loss},  \\
        MixUp~\cite{MixUp_Loss},    \\
        RICAP~\cite{Ricap_Loss}     \\
    \end{tabular}   \\
  
    \bottomrule
  \end{tabular}

\end{table}

\begin{table}[H]
\small
\centering
\caption{Network architectures and names of the models pre-trained on ImageNet.}
\label{table:models_xtics_imagenet}
\begin{tabular}{p{1.7cm}l}
    \toprule
    \begin{tabular}{c}
         Network \\
         Architectures 
    \end{tabular} &
    \hspace{2em}
    \begin{tabular}{l}
        VGG~\cite{VGG}, DenseNet~\cite{DenseNet},     \\
        ResNet~\cite{ResNet},    ResNeXt~\cite{ResNeXt},      \\
        Inception-ResNet~\cite{InceptionResNet_2017},    EfficientNet~\cite{EfficientNet_rethinking_model_scaling_2019},      \\ DPN~\cite{DPN_2017},  SEResNet~\cite{SEResNet_squeeze_and_excitation_2020},   \\
        HRNet~\cite{HRNet_deep_high_resolution_2021},   DeiT~\cite{deit_training_data_efficient_2021}     \\ PNasNet~\cite{PNasNet_progressive_neural_architecture_2017},   Visformer~\cite{visformer_the_vision_friendly_2021},       \\
        Vision Transformers (ViT)~\cite{ViT_an_image_is_2020}
    \end{tabular} \\ 
    
    \midrule
    
    \begin{tabular}{c}
         Pre-trained \\
         Models \\\cite{Timm_Models_Repo}
    \end{tabular} &
    \hspace{2em}
    \begin{tabular}{l}

         densenet161, deit\_tiny\_distilled\_patch16\_224, dpn68b, efficientnet\_b2, dpn92,  \\
         gluon\_resnet50\_v1d, gluon\_seresnext50\_32x4d, gluon\_seresnext101\_32x4d,   \\
         hrnet\_w18\_small\_v2, ig\_resnext101\_32x8d, ig\_resnext101\_32x16d,      \\
         legacy\_seresnet50, ig\_resnext101\_32x32d, ig\_resnext101\_32x48d,   \\
         inception\_resnet\_v2, pnasnet5large, resnet18, seresnext50\_32x4d,     \\
         tf\_efficientnet\_b5, tf\_efficientnet\_l2\_ns, vgg13, visformer\_small,     \\
         vit\_base\_patch16\_224, vit\_tiny\_patch16\_384
    \end{tabular}   \\
    
    \bottomrule
  \end{tabular}

\end{table}

\subsection{Matching by Predicted Probabilities Only}
\label{appendix_description_second_matching_criterion}
\begin{algorithm}[H]
\caption{Strategy for matching two datasets based on the predicted model probabilities only.}
\label{algorithm_second_matching_strategy}
\begin{algorithmic}
    \Procedure {MatchPredictions}{$f_{\bm{\theta}}$, $\mathcal{D_{\mathrm{src}}}$, $\mathcal{D_{\mathrm{tgt}}}$, $\epsilon$}
    \State $\mathcal{P}_{\mathrm{src}} \leftarrow \Call{Predict}{f_{\bm{\theta}}, \mathcal{D_{\mathrm{src}}}}$
    \State $\mathcal{P}_{\mathrm{tgt}} \leftarrow \Call{Predict}{f_{\bm{\theta}}, \mathcal{D_{\mathrm{tgt}}}}$ 
    \vspace{10pt}
    \State Initialize 3 empty arrays:
    \State $\mathcal{P}_{\mathrm{src\_matched}}$ = [ ]
    \State $\mathcal{P}_{\mathrm{tgt\_matched}}$ = [ ]
    \State $\mathcal{P}_{\mathrm{tgt\_unmatched}}$ = [ ]
    \vspace{10pt}
    \State $N \leftarrow \Call{Length}{\mathcal{P}_{\mathrm{tgt}}}$
    \vspace{10pt}
        
    \For{$i\gets 1, N$}
        \State $y_{t}, \hat{y}_{t}, \hat{p}_{t} \leftarrow \mathcal{P}_{\mathrm{tgt}}[i]$
        \State $y_{s}, \hat{y}_{s}, \hat{p}_{s}, is\_match \leftarrow \Call{Match}{y_{t}, \hat{y}_{t}, \hat{p}_{t},\mathcal{P}_{\mathrm{src}}, \epsilon}$
        \If{$is\_match$}
            \State $\mathcal{P}_{\mathrm{src\_matched}}.append((y_{s}, \hat{y}_{s}, \hat{p}_{s}))$
            \State $\mathcal{P}_{\mathrm{tgt\_matched}}.append((y_{t}, \hat{y}_{t}, \hat{p}_{t}))$
        \Else
            \State $\mathcal{P}_{\mathrm{tgt\_unmatched}}.append((y_{t}, \hat{y}_{t}, \hat{p}_{t}))$
        \EndIf
    \EndFor
        
    \Return {$\mathcal{P}_{\mathrm{src\_matched}}, \mathcal{P}_{\mathrm{tgt\_matched}}, \mathcal{P}_{\mathrm{tgt\_unmatched}}$}
    \EndProcedure 
    \newline
        
    \vspace{10pt}
    \Function{Match}{$y_{t}, \hat{y}_{t}, \hat{p}_{t}, \mathcal{P}_{\mathrm{src}}, \epsilon$}
    \State Initialize an empty $\mathcal{P}_{\mathrm{matched}}$ = [ ] 
    \ForAll {$(y, \hat{y}, \hat{p}) \in \mathcal{P}_{\mathrm{src}}$}
        
        \If{$\hat{p}_{t} \in [\hat{p} - \epsilon, \hat{p} + \epsilon]$}
            \State $\mathcal{P}_{\mathrm{matched}}.append((y, \hat{y}, \hat{p}))$
        \EndIf
    \EndFor
    \If{|$\mathcal{P}_{\mathrm{matched}}$| > 0}
        \State $(y_{s}, \hat{y}_{s}, \hat{p}_{s}) \gets \Call{RandomSelect}{\mathcal{P}_{\mathrm{matched}}}$
        \setlength\parindent{24pt} 
        \State $\mathcal{P}_{\mathrm{src}}.remove((y, \hat{y}, \hat{p}))$  
        \State \Return $(y_{s}, \hat{y}_{s}, \hat{p}_{s}, is\_match)$
    \Else 
        \State \Return $(0, 0, 0, False)$
    \EndIf
    \EndFunction
    \newline
    \Comment The \Call{Predict}{.} function returns the predictions of a model in an array that consists of tuples $(y, \hat{y}, \hat{p})$, where $y$, $\hat{y}$, and $\hat{p}$ denote the ground truth label, the predicted label, and the predicted probability, respectively.
    \newline 
    \Comment The \Call{RandomSelect}{.} function returns a randomly selected tuple $(y, \hat{y}, \hat{p})$ from the array that contains the input tuples.
    \newline 
    \Comment The $\mathcal{P}_{\mathrm{src}}.remove((y, \hat{y}, \hat{p}))$ in \Call{RandomSelect}{.} accounts for random selection without replacement.
    \end{algorithmic}
\end{algorithm}

\subsection{Results for Matching by Predicted Probabilities Only}
\label{appendix_results_from_matching_by_only_the_predicted_probabilities}



\begin{figure}[!htb]
\subfloat[\label{fig:acc_matched_acc_cifar10_vs_cifar10.1_2nd_match_criterion}]{%
  \includegraphics[ width=0.48\linewidth]{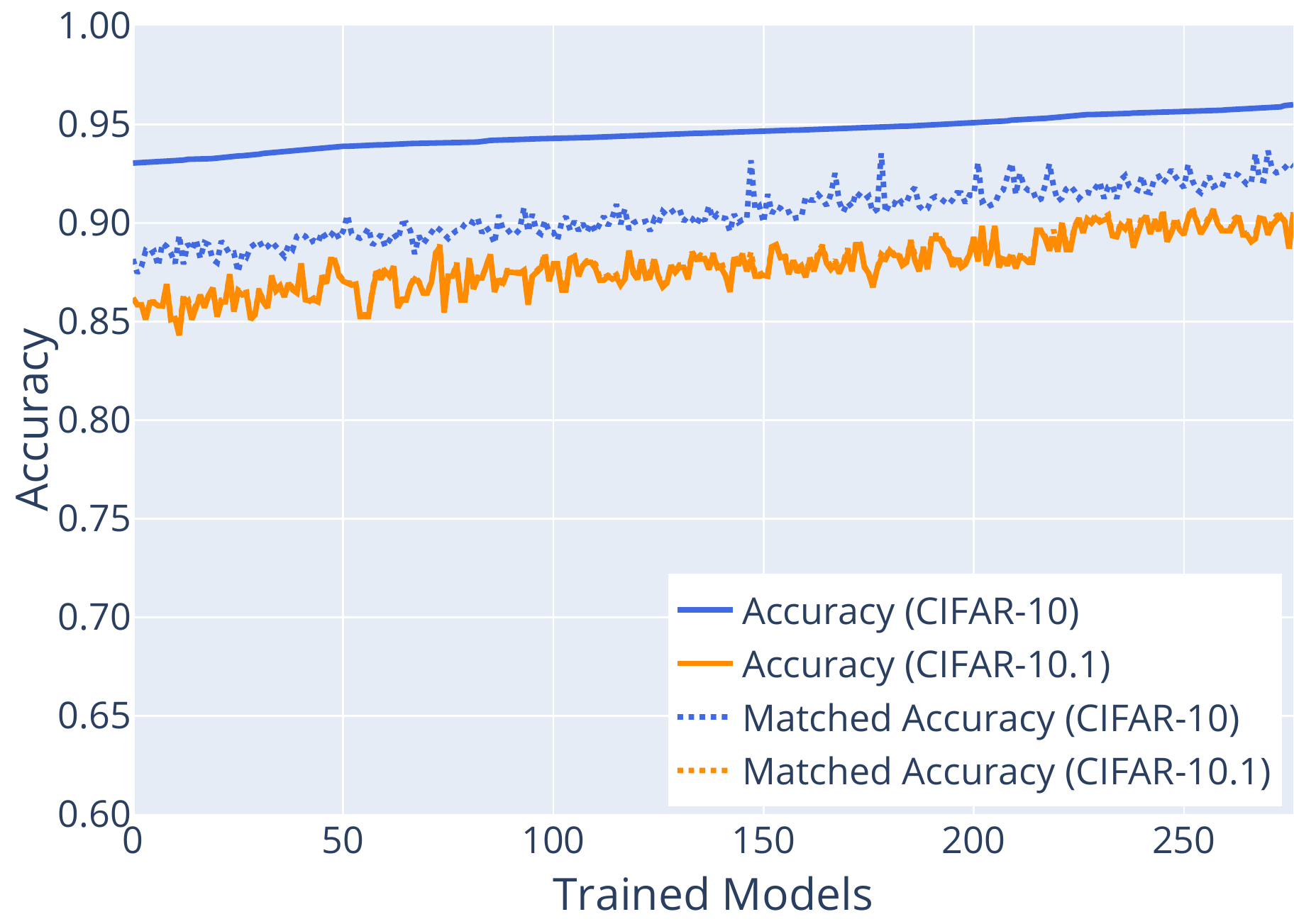}}\hfill
\subfloat[\label{fig:xtics_cifar10_vs_cifar10.1_2nd_match_criterion}]{%
  \includegraphics[ width=0.48\linewidth]{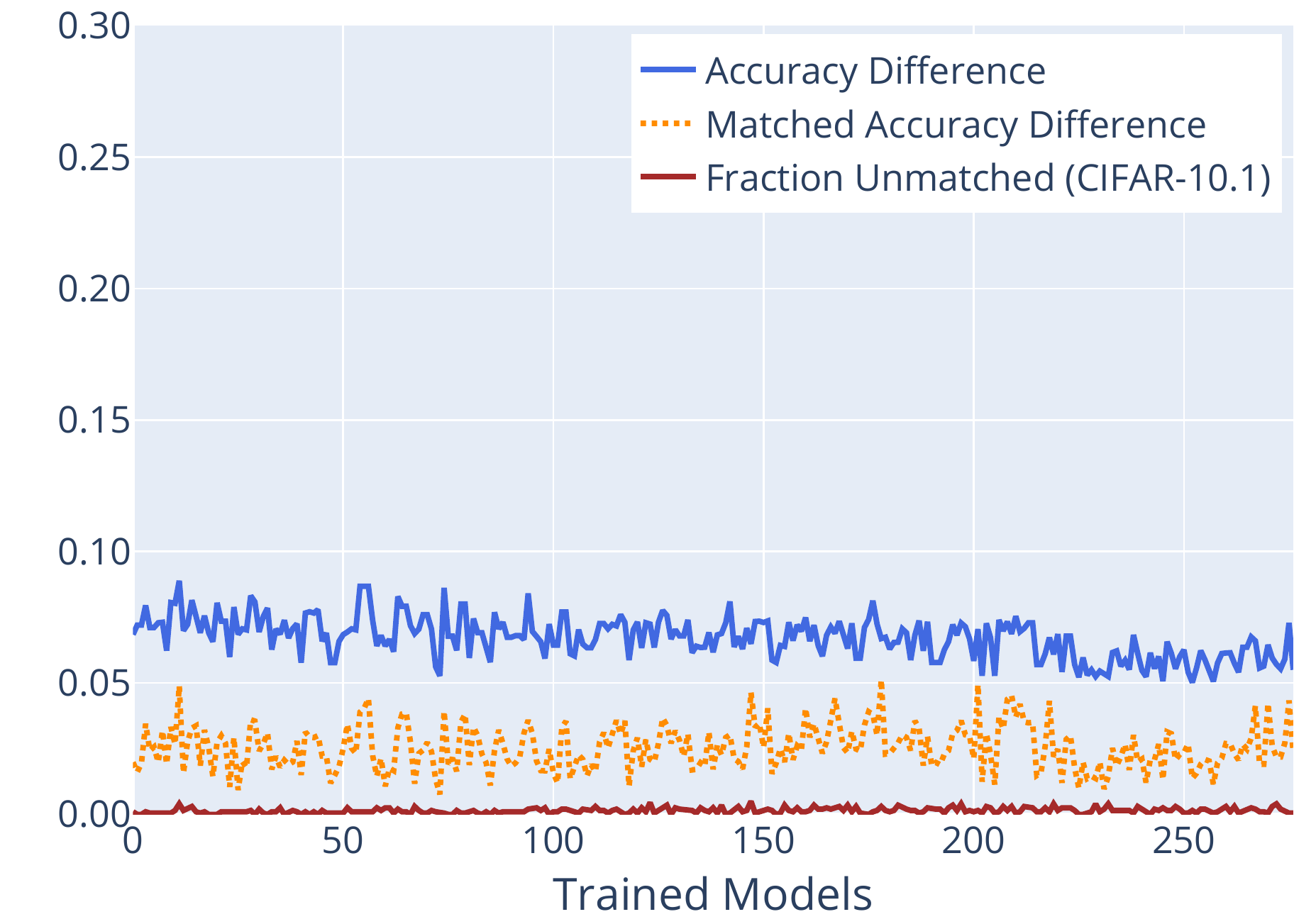}}\vfill
\subfloat[\label{fig:acc_matched_acc_cifar10_vs_cifar10.2_2nd_match_criterion}]{%
  \includegraphics[ width=0.48\linewidth]{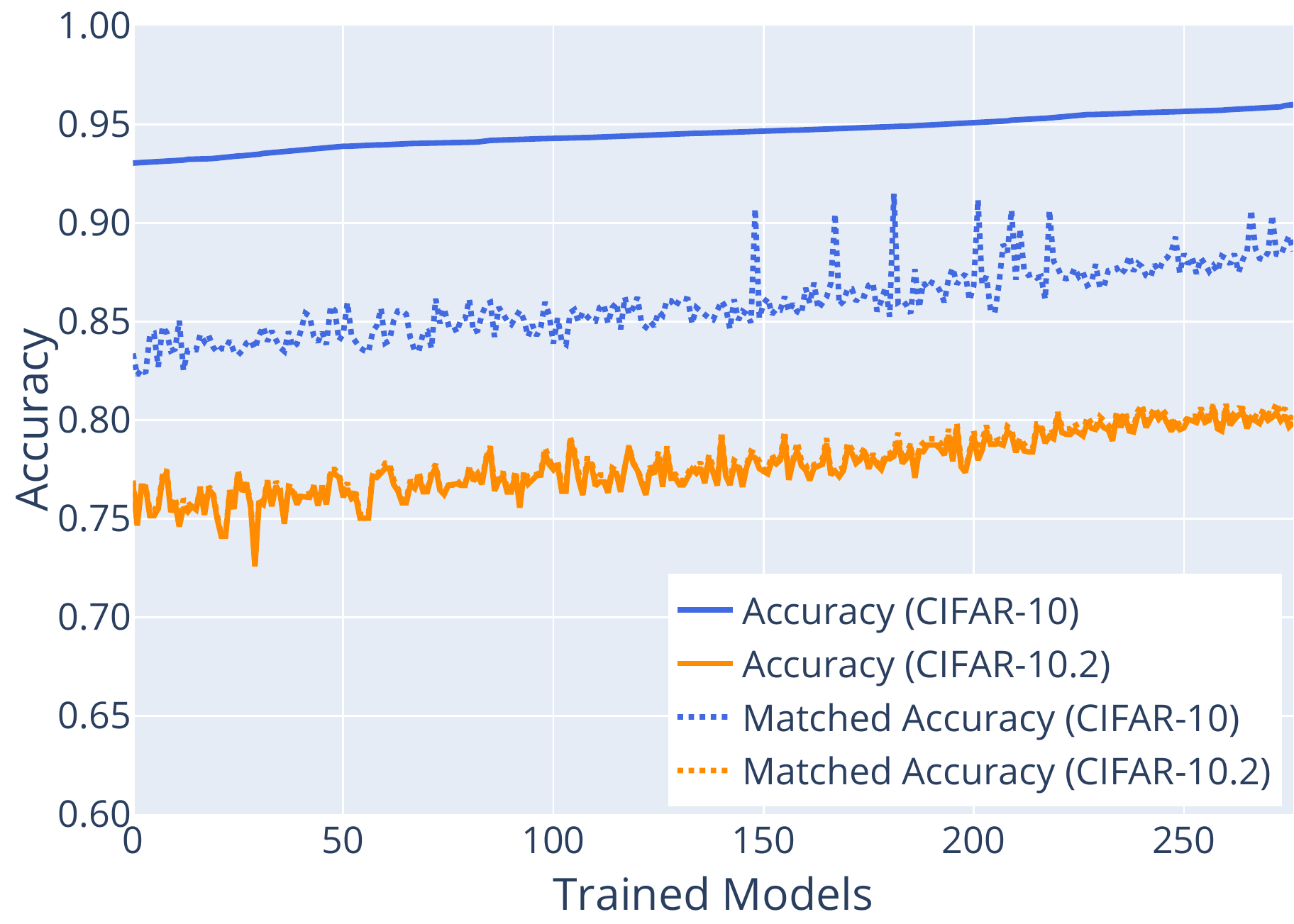}}\hfill
\subfloat[\label{fig:xtics_cifar10_vs_cifar10.2_2nd_match_criterion}]{%
  \includegraphics[ width=0.48\linewidth]{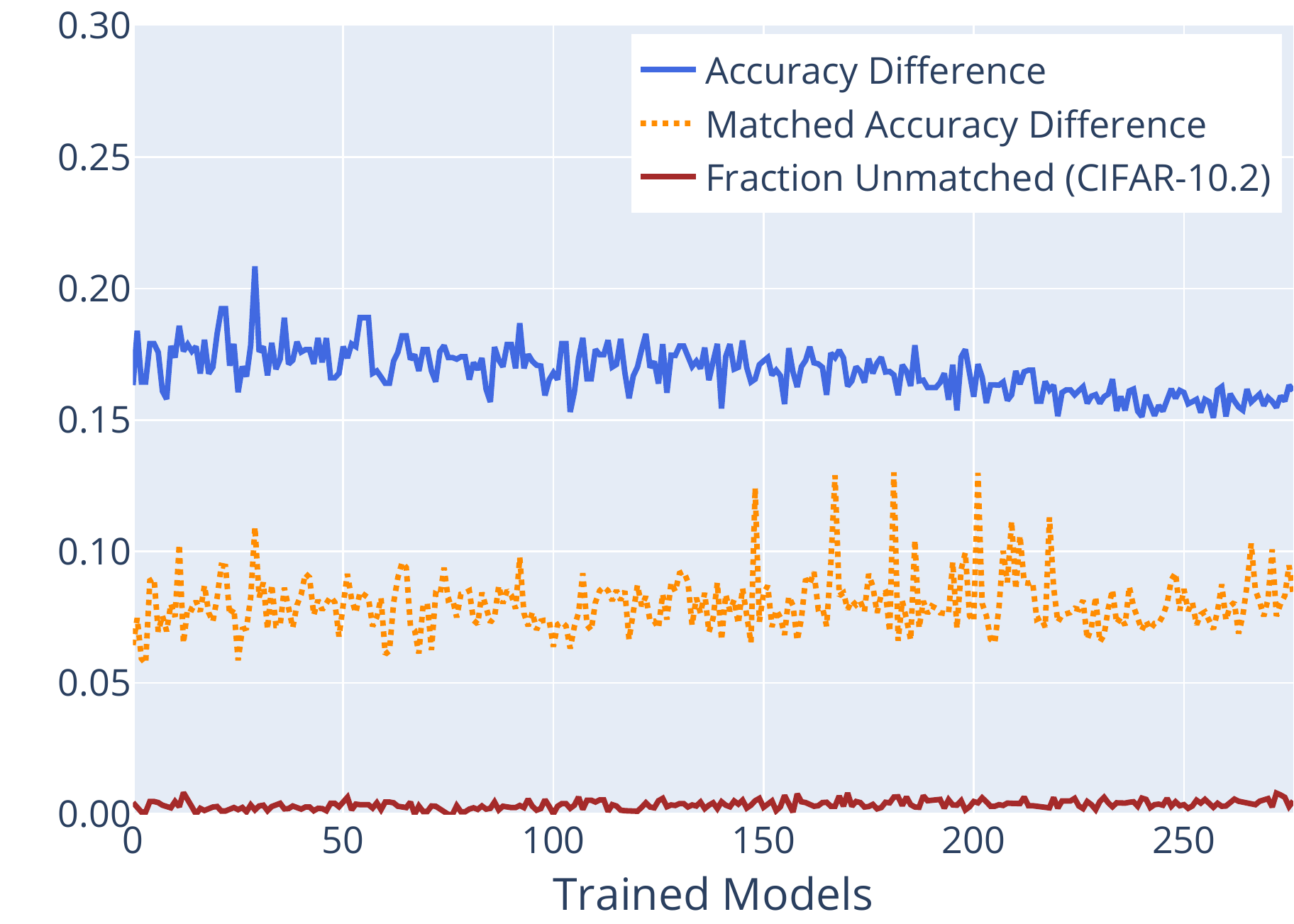}}\vfill
  \subfloat[\label{fig:acc_matched_acc_cifar10_vs_cinic10_2nd_match_criterion}]{%
  \includegraphics[ width=0.48\linewidth]{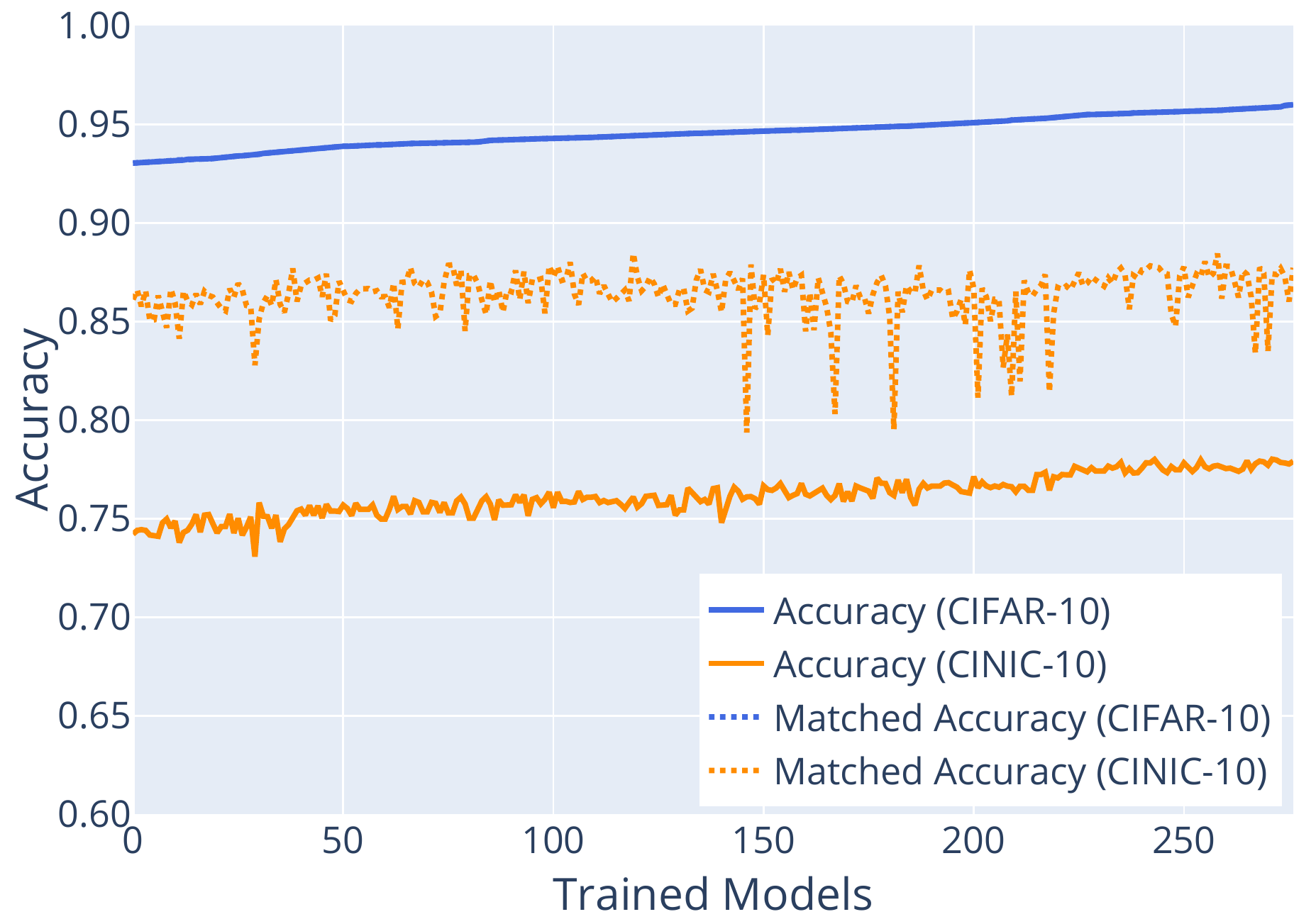}}\hfill
\subfloat[\label{fig:xtics_cifar10_vs_cinic10_2nd_match_criterion}]{%
  \includegraphics[ width=0.48\linewidth]{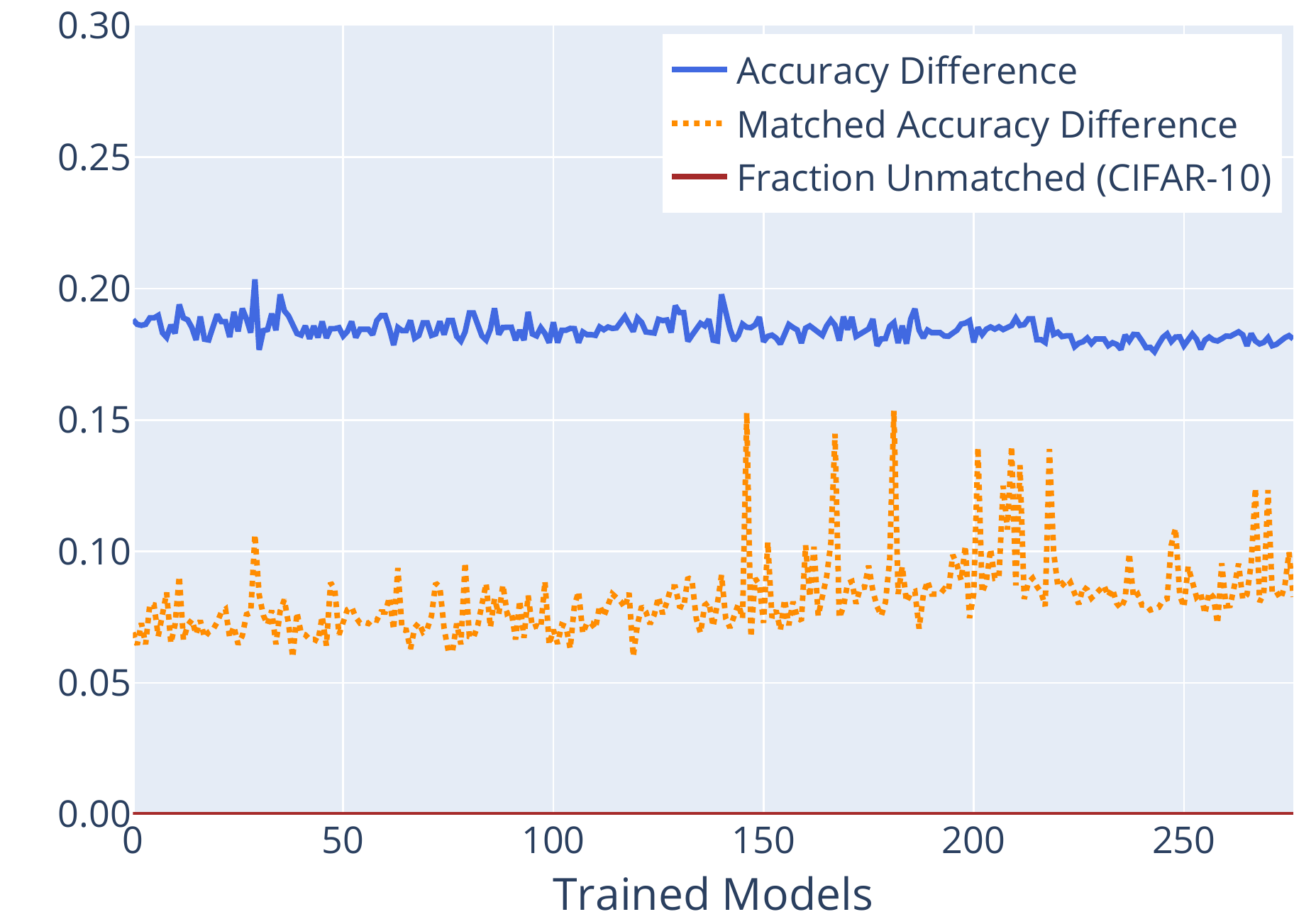}}\vfill

\caption{Plots for accuracy and matched accuracy, when matched by predicted probabilities only for the following dataset pairs: (a) CIFAR-10 versus CIFAR-10.1, (c) CIFAR-10 versus CIFAR-10.2, and (e) CIFAR-10 versus CINIC-10. Plots for accuracy difference, matched accuracy difference, and fraction unmatched for (b) CIFAR-10 versus CIFAR-10.1, (d)  CIFAR-10 versus CIFAR-10.2, and (f) CIFAR-10 versus CINIC-10. The models in all plots are sorted according to increasing accuracy on CIFAR-10.}
    \label{fig:acc_matched_acc_xtics_cifar_related_2nd_match_criterion}
\end{figure}

\begin{figure}[!htb]
\subfloat[\label{fig:acc_matched_acc_imagenet_vs_imagenetv2_2nd_match_criterion}]{%
  \includegraphics[ width=0.48\linewidth]{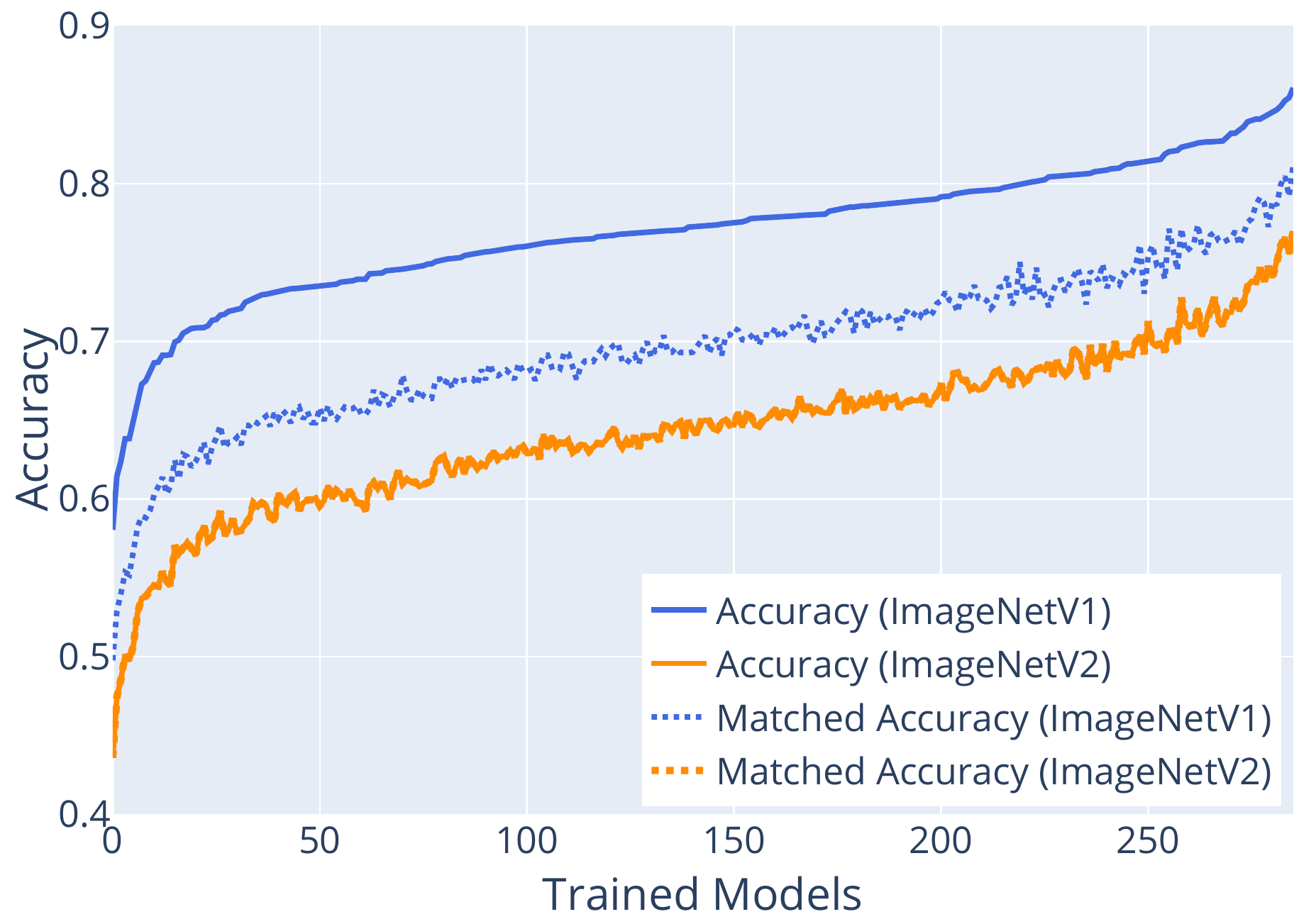}}\hfill
\subfloat[\label{fig:xtics_imagenet_vs_imagenetv2_2nd_match_criterion}]{%
  \includegraphics[ width=0.48\linewidth]{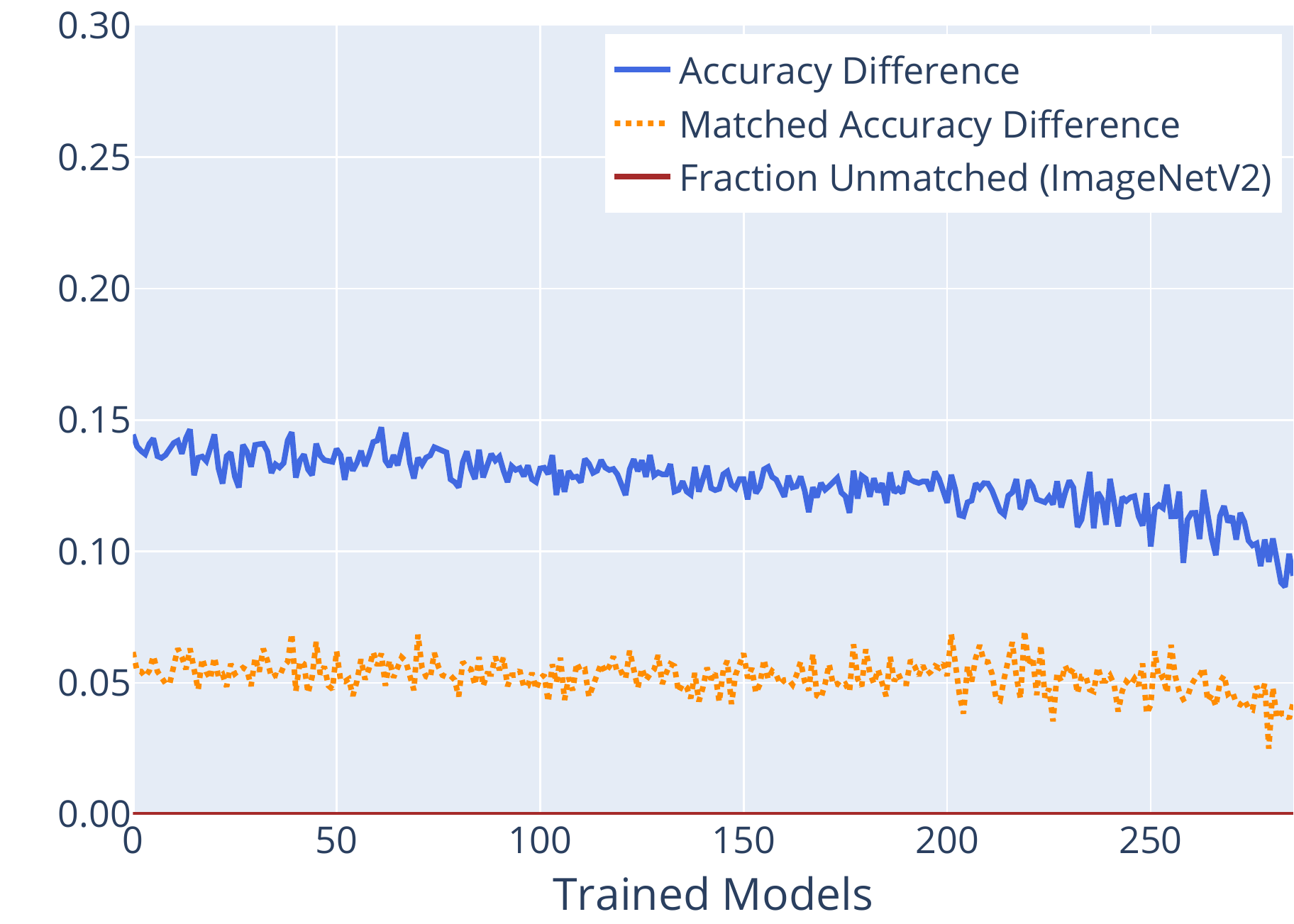}}

\caption{(a) Accuracy and matched accuracy plots and (b) difference in accuracy and matched accuracy for all models for the ImageNet and ImageNetV2 dataset pair. The models on all plots are sorted according to increasing accuracy on ImageNetV1.}
    \label{fig:acc_matched_acc_xtics_imagenet_related_2nd_match_criterion}
\end{figure}

\clearpage

\subsection{Additional Results for the Accuracy versus Confidence Curves}
\label{appendix_additional_examples_cal_curves_and_density_plots}


\begin{figure}[!htb]
\subfloat[\label{fig:_swsl_resnet18cal_curves}]{%
  \includegraphics[width=0.47\linewidth]{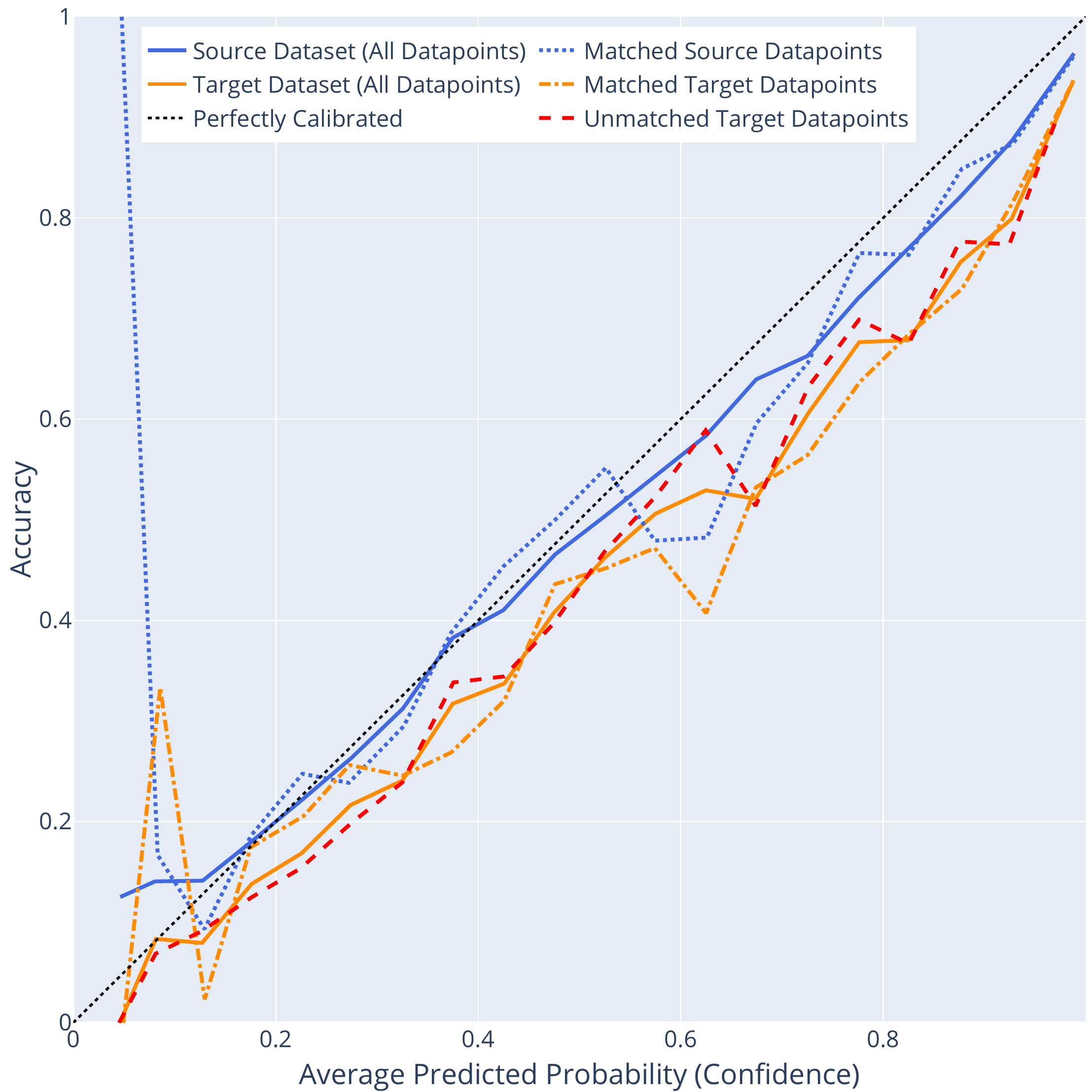}}\hfill
\subfloat[\label{fig:_swsl_resnet18_densities}]{%
  \includegraphics[width=0.51\linewidth]{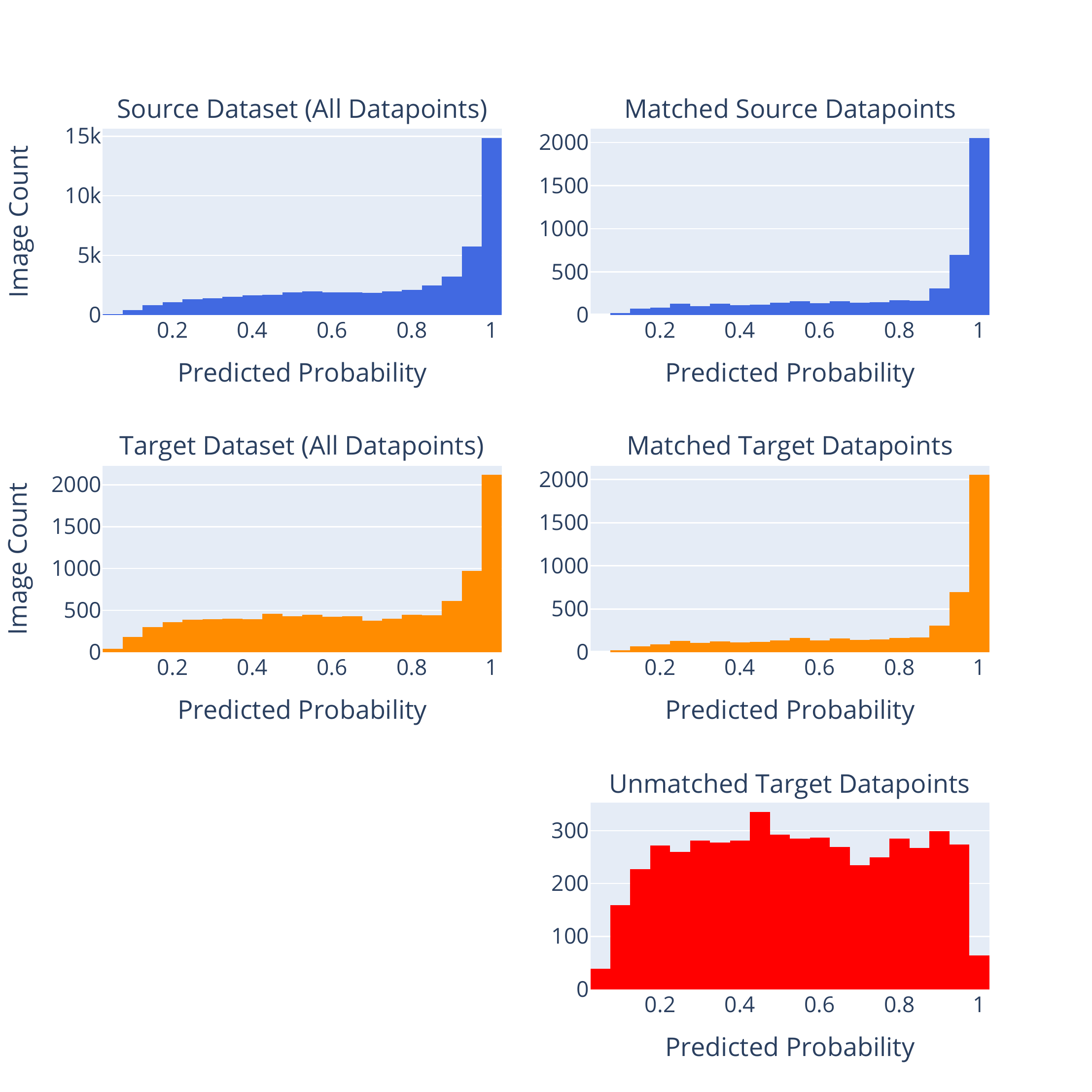}}

\caption{The plots of two characteristic of a pre-trained ResNet18 (swsl\_resnet18) model~\cite{Timm_Models_Repo} when evaluated on various subsets of the ImageNetV1 and ImageNetV2 datasets, as generated by our proposed evaluation protocol: (a) Accuracy versus Confidence and (b) Density plots for predicted probabilities.}
    \label{fig:cal_curves_and_density_plots_swsl_resnet18}
\end{figure}

\end{document}